\pdfoutput=1

\documentclass[11pt]{article}

\usepackage[preprint]{mtsummit25}
\usepackage{copyright}

\usepackage{times}
\usepackage{latexsym}

\usepackage[T1]{fontenc}

\usepackage[utf8]{inputenc}

\usepackage{microtype}

\usepackage{inconsolata}

\usepackage{graphicx}

\usepackage{amsmath,amssymb}
\usepackage{algorithm2e}
\usepackage{multirow}
\usepackage{booktabs}
\usepackage[colorinlistoftodos]{todonotes}
\usepackage{cleveref}
\usepackage{xspace}
\usepackage{floatrow}
\usepackage{subcaption}
\usepackage{caption}

\newcommand{\model}[1]{\textsc{#1}}
\newcommand{\corpus}[1]{\textbf{#1}}
\newcommand{\testset}[1]{\textbf{#1}}

\newcommand{\zpTodo}[1]{\todo[disable,color=cyan!30]{#1}}
\newcommand{\done}[1]{\todo[disable]{#1}}
\newcommand{\edite}[1]{#1}
\newcommand{\rebuttal}[1]{#1}

\newcommand{\pvalue}[1]{\textcolor{olive}{#1}}

\newcommand{\posUnif}{\model{unifPE}\xspace}
\newcommand{\tedI}{\corpus{TED-U}\xspace}
\newcommand{\tedII}{\corpus{TED-G}\xspace}

\newcommand{\FTnllbI}{\model{FT-NLLB-U}\xspace}
\newcommand{\FTnllbII}{\model{FT-NLLB-G}\xspace}
\newcommand{\FTtowerI}{\model{FT-Tower-U}\xspace}
\newcommand{\FTtowerII}{\model{FT-Tower-G}\xspace}

\newcommand{\UNIFnllbI}{\model{Unif-NLLB-U}\xspace}
\newcommand{\UNIFnllbII}{\model{Unif-NLLB-G}\xspace}
\newcommand{\UNIFtowerI}{\model{Unif-Tower-U}\xspace}
\newcommand{\UNIFtowerII}{\model{Unif-Tower-G}\xspace}

\newcommand{\SHAPEnllbI}{\model{SHAPE-NLLB-U}\xspace}
\newcommand{\SHAPEnllbII}{\model{SHAPE-NLLB-G}\xspace}
\newcommand{\SHAPEtowerI}{\model{SHAPE-Tower-U}\xspace}
\newcommand{\SHAPEtowerII}{\model{SHAPE-Tower-G}\xspace}

\title{Investigating Length Issues in Document-level Machine Translation}

\author{
 \textbf{Ziqian Peng\textsuperscript{1,2}},
 \textbf{Rachel Bawden\textsuperscript{2}},
 \textbf{François Yvon\textsuperscript{1}},
\\
\\
 \textsuperscript{1}Sorbonne Université \& CNRS, ISIR, Paris, France,
 \\
 \textsuperscript{2}Inria, Paris, France,
\\
    \texttt{\{ziqian.peng,francois.yvon\}@isir.upmc.fr}~~~~
    \texttt{rachel.bawden@inria.fr}
}

\begin{document}
\maketitle
\begin{abstract}
Transformer architectures are increasingly effective at processing and generating very long chunks of texts, opening new perspectives for document-level machine translation (MT). In this work, we challenge the ability of MT systems to handle texts comprising up to several thousands of tokens. We design and implement a new approach designed to precisely measure the effect of length increments on MT outputs. Our experiments with two representative architectures unambiguously show that (a)~translation performance decreases with the length of the input text; (b)~the position of sentences within the document matters, and translation quality is higher for sentences occurring earlier in a document. We further show that manipulating the distribution of document lengths and of positional embeddings only marginally mitigates such problems. Our results suggest that even though document-level MT is computationally feasible, it does not yet match the performance of sentence-based MT.   
\end{abstract}

\section{Introduction \label{sec:intro}}
Statistical and neural machine translation (MT) architectures \citep{koehn2020neural} have been designed to process isolated sentences, limiting their ability to properly handle discourse phenomena, such as coherence and cohesion, the modelling of which requires longer contexts \citep{fernandes-etal-2023-translation}. A first step to address this shortcoming has been to augment the source and/or the target side with a couple of preceding sentences \citep{tiedemann-scherrer-2017-neural}. Multiple approaches to encode and fully exploit such extended contexts have been proposed \citep{popescubelis-2019-context,maruf-etal-2021-survey,castilho-knowles-2024-asurvey}\done{cite more, the survey in 2024?} and have been shown to improve the ability of MT engines to preserve local discourse coherence and cohesiveness through word-sense disambiguation or the resolution of anaphoric references \citep{bawden-etal-2018-evaluating,voita-etal-2018-context}. Most of these approaches continue to process texts on a per-sentence basis with an extended context, even though attempts have also been made to process continuous chunks of texts comprising several sentences \citep{scherrer-etal-2019-analysing, lopes-etal-2020-document,ma-etal-2020-simple,ma-etal-2021-comparison,lupo-etal-2022-divide,wu-etal-2023-document}.\done{cite more?}

The ability of today's neural MT models---relying on encoder-decoder or decoder-only architectures---to handle large context lengths, up to thousands of tokens \citep{peng2024yarn}, opens new perspectives to go beyond \emph{context-augmented MT} and develop fully-fledged \emph{document-level MT}, where the entire document context is available at once, and where the target text is generated in a single pass.\footnote{These perspectives are, for instance, explored in the latest edition of the WMT shared task on General Machine Translation \citep{kocmi-etal-2024-findings}.} Two main technical novelties have made this possible: (a)~more efficient computation in the attention layers \citep{tay-etal-2022-efficient} and (b)~changes in the design of positional encodings (PEs). In particular, replacing the sinusoidal absolute PEs \edite{(APEs)} of \citep{vaswani-etal-2017-attention} with methods like ALIBI \cite{press2022train} and RoPE \citep{SU2024127063}, which lend themselves well to length extrapolation \citep{sun-etal-2023-length,zhao-etal-2024-length}, seems to make today's transformers amenable to the processing of arbitrarily long contexts \citep{mohtashami2023landmark,han-etal-2024-lm}.

In this work, we challenge the ability of contemporary MT models to effectively handle long spans of texts. For this, we develop a new methodology for assessing the impact of length variations on MT performance. We perform a series of controlled experiments with two representative neural MT systems, where the same documents are processed by chunks of increasing lengths in a document-level manner and show that (a)~MT performance tends to degrade with the length of the source document, (b)~length issues happen even for in-distribution lengths and get worse when extrapolating to unseen document lengths, and (c)~most of the degradation happens in the final parts of the translation. 
Hypothesising that this may be due to a mismatch between the distribution of train and test PEs, we explore a possible mitigation, which flattens the distribution of PEs during training. We observe a consistent improvement of automatic metric scores for the \edite{APE-based} vanilla encoder-decoder model, while the \edite{RoPE based} decoder-only model remains mostly unaffected. 
In summary, our main contributions are: (a)~a new approach to the detection and diagnosis of length issues in document-level MT, (b)~a new variant of the \model{SHAPE} \cite{kiyono-etal-2021-shape} method, which improves the distribution of PEs during training, and (c)~a confirmation that  (perhaps for lack of an appropriate document-level evaluation tool) sentence-level MT remains a strong baseline in most settings.

\section{Related work \label{sec:related}}
\subsection{Document-level MT \label{ssec:related-dlmt}}
Previous attempts to incorporate more contextual information in MT models can be roughly categorized into two categories: context-augmented MT, also called \emph{Doc2Sent} in \citep{sun-etal-2022-rethinking}, and document-level MT, also called \emph{Doc2Doc}. Recent surveys of this field include \citep{popescubelis-2019-context,maruf-etal-2021-survey,castilho-knowles-2024-asurvey}.

Translation of discourse phenomena, such as lexical consistency, reference, and word sense disambiguation, requires inter-sentential context \citep{bawden-etal-2018-evaluating,wong-etal-2020-contextual,fernandes-etal-2023-translation}. This has motivated the integration of extended (local) contexts in \emph{Doc2Sent} models.
Such approaches include concatenation-based methods \citep{tiedemann-scherrer-2017-neural}; architecture adaptations to process context in different components of the same encoder \citep{ma-etal-2020-simple,wu-etal-2023-document}, in a dedicated encoder \citep{voita-etal-2018-context, zhang-etal-2018-improving}, or via hierarchical attention networks \citep{miculicich-etal-2018-document, maruf-etal-2019-selective, yin-etal-2021-context}; cache-based methods using a short-term MT memory
\citep{maruf-haffari-2018-document, tu-etal-2018-learning, yang-etal-2019-enhancing,dobreva-etal-2020-document} 
and multi-pass decoding algorithms
\citep{voita-etal-2019-context,yu-etal-2020-better,kang-etal-2020-dynamic}.

Translating sentence by sentence, even with augmented contexts, still fails to capture phenomena related to coherence and consistency  \citep{fernandes-etal-2023-translation}, motivating \emph{Doc2Doc} approaches to process documents as a whole.
This can be done with concatenation-based methods \citep{tiedemann-scherrer-2017-neural,sun-etal-2022-rethinking,karpinska-iyyer-2023-large}, \edite{along with} sliding window attention \citep{zhuocheng-etal-2023-addressing,liu-etal-2023-only} and group attention \citep{bao-etal-2021-g} to address the issue of quadratic complexity. Other strategies include focusing on improving training through data augmentation with a balanced length distribution \citep{sun-etal-2022-rethinking} and richer context-dependent phenomena \citep{lupo-etal-2022-divide,wu-etal-2024-importance}, or on better training strategies with multilingual denoising pretraining \citep{lee-etal-2022-docmt5}, adapted loss functions \citep{lupo-etal-2022-focused}, and enriched positional encodings \citep{li-etal-2023-ptransformer,lupo-etal-2023-encoding}.
Multiple \edite{methods} 
have recently emerged \edite{for large language models (LLMs)} \citep{wang-etal-2023-document-level}, which also show a decline in translation quality as input length increases \citep{wang-etal-2024-benchmarking}.\footnote{They also confirmed the effectiveness of training LLMs on documents of varied sizes (similar to \citet{sun-etal-2022-rethinking}).} These include a two-stage training recipe with the use of a monolingual corpus and high-quality parallel documents \citep{xu-2024-paradigm,alves2024tower}, and applying LLMs as post-editors \citep{koneru-etal-2024-contextual}. 

\subsection{Extrapolating PEs \label{ssec:related-pe}}
\done{Introduce APE / RPE}
Since self-attention is position-agnostic, 
PEs are used to provide position information in Transformer models. PEs embed the absolute token position (APEs) \citep{vaswani-etal-2017-attention}, or the relative distance between tokens (RPEs) \citep{shaw-etal-2018-self,raffel2023exploring,press2022train}, with RoPE \citep{SU2024127063} being the go-to approach in recent LLMs such as Llama2 \citep{touvron2023llama}.
Despite RPEs yielding better length extrapolation ability than APEs, both of them struggle to efficiently extrapolate input lengths beyond the predefined maximum training length 
\citep{dai-etal-2019-transformerxl,chen2023extending,peng2024yarn,zhao-etal-2024-length}, motivating the development of input extension methods for PEs. 

For APEs, 
\model{SHAPE} \citep{kiyono-etal-2021-shape} offsets all indices in a sequence by some random values. Its authors show that this simple technique mimics the computation of RPEs at a much smaller cost and helps to improve the interpolation abilities of a vanilla encoder-decoder model, as measured by BLEU \citep{papineni-etal-2002-bleu} with long pseudo-documents. Our experiments confirm that this technique is effective using actual document contexts and a sounder experimental methodology, based on paired tests, and using COMET \citep{rei-etal-2020-comet}.\done{Read this one !}     
\citeauthor{sinha-etal-2022-curious}'s (\citeyear{sinha-etal-2022-curious}) experiments adopt a setting similar to ours, offsetting the absolute value of APEs' input to evaluate their ability to capture relative distances between tokens. Their results, like ours, illustrate the lack of robustness of APEs and suggest that they overfit their training data.  

For RPEs, especially RoPE, both position interpolation (PI) and position extrapolation methods have been proposed.
PI methods interpolate positions to extrapolate context length directly during inference or through fine-tuning \citep{chen2023extending,peng2024yarn}.
The position extrapolation methods aim to extend context using documents that are shorter than the predefined maximum length. For example,
RandPos \citep{ruoss-etal-2023-randomized} randomly maps position indices to a much larger interval with the original word order, and
PoSE \citep{zhu2024pose} divides each training sequence into $N$ chunks and adjusts the position indices of every chunk except the first one by adding a uniformly sampled offset, within the scope of a predefined maximal length.

\section{Methods and Metrics \label{sec:methods}}
\subsection{Holistic Document-Level MT \label{ssec:challenge}}
\done{take some part from taln2024-long-doc}
\done{introduction for "such questions" mentioned below}
\done{Explain why this is interesting - for instance WMT 24 - full context setting}
Compared to sentence-based MT, holistic document-Level MT (Doc2Doc) possesses several appealing features, as it gives access to all the available textual context. This should enable the MT system to improve on global aspects pertaining for instance to coherence and cohesion.\done{Add refs.}
However, Doc2Doc also introduces several new challenges compared to the
Sent2Sent scenario:
\begin{enumerate}
 \setlength\itemsep{0em}
     \item in Doc2Doc, input texts are longer, causing a computational overhead due to the quadratic complexity of attention \citep{tay-etal-2022-efficient}.
    \item for longer inputs, attention weights are spread over a larger number of tokens \citep{herold-ney-2023-improving}; however,
    at each decoding step, most attention needs to remain concentrated on the corresponding local source context \citep{bao-etal-2021-g}. This is in contrast with Doc2Sent, where sentence alignment is readily available.
    \item decoding longer sequences increases the impact of search errors and of exposure bias \citep{ranzato-etal-2016-sequence}. Beam search also becomes more difficult due to the input length.
    \item output sentences may not always stand in one-to-one correspondence with source sentences, which complicates the computation of automatic metrics, which are designed to evaluate one-to-one mappings between hypotheses and references.
\end{enumerate}
These differences motivate our main research questions, which we rephrase as: (a)~For existing models, does \emph{Doc2Doc} bring more benefits than disadvantages compared with \emph{Sent2Sent}? (b)~How do these results vary with the input document length? (c)~Which methods and metrics can we use to automatically evaluate the impact of length differences? 

\subsection{Shades of BLEU \label{ssec:bleus}}
Answering such questions requires metrics for comparing holistic translations with sentence-based translations: as the number of segments produced by the former may differ from the number of source segments, a basic requirement is that they allow the evaluation of translation hypotheses with more (or fewer) sentences than the source (for quality estimation scores) and/or the reference (for reference-based metrics). However, most existing document-level MT approaches still rely on BLEU \cite{papineni-etal-2002-bleu}, despite its well-documented shortcomings \citep{callison-burch-etal-2006-evaluating,reiter-2018-structured,mathur-etal-2020-tangled,dahan-etal-2024-evaluating-dlmt}; or rather a variant dubbed \emph{d-BLEU} by \citet{liu-etal-2020-multilingual}.\footnote{\citet{hendy2023good} also consider a variant of COMET \cite{rei-etal-2022-comet} while \citet{zhuocheng-etal-2023-addressing} introduce d-ChrF, a document-level version of ChrF \citep{popovic-2015-chrf}.} We accordingly focus on BLEU in this section, noting that the same questions would need to be addressed with any metric relying on sentence-based surface comparison (e.g., METEOR \cite{banerjee-lavie-2005-meteor}, TER \cite{snover-etal-2006-study}, BertSCORE \citep{zhang-etal-2020-bertscore}, PRISM \citep{thompson-post-2020-automatic}, COMET \citep{rei-etal-2020-comet}, and many others).\footnote{We choose to evaluate using the standard metrics, BLEU and COMET, rather than evaluation approaches specifically designed to test the use of increased context. This choice is motivated by the fact that the score differences we observe reveal a significant degradation in translation quality for longer documents, indicating greater problems than those targeted by finer-grained evaluation techniques.}

BLEU is computed by counting, sentence by sentence, the number of $n$-grams (for $n \in [1\!:\!4]$) shared by each translation hypothesis and its human reference. These counts are aggregated and turned into frequencies, then averaged (geometrically) at the corpus level. Finally, a length penalty is applied to degrade the score when the cumulated length of the hypotheses is shorter than that of the references. BLEU is a corpus-level score that depends on sentence alignments. d-BLEU is also a global score but counts common $n$-grams at the document level. As a consequence, d-BLEU, which records matches for larger spans than BLEU, delivers higher scores, as the opportunities to match $n$-grams are greater for a wider window.\footnote{This effect is well known, e.g.\ in \citep[Figure~7]{koehn-knowles-2017-six}, where BLEU increases when considering sentence groups of increasing lengths (at least for a certain length range), where we would expect a decrease, as the length is often linked to syntactic complexity and therefore to translation difficulty. We reproduce this observation in Figure~\ref{fig:bleu}.} 
These two scores cannot be compared, and we contend that their shortcomings make them inappropriate for analysing length-related issues in MT.

\edite{An alternative to d-BLEU is to perform evaluation at the document level, rather than the corpus level. This can be implemented either as (a)~calculating one BLEU score (with realignment) per document, then averaging at the corpus level or (b)~calculating the equivalent of sentence-level BLEU scores \citep{lin-och-2004-automatic} but where each segment is a concatenated full document rather than a sentence.
However, (a) counts matches at the sentence-level, which requires a realignment between translated and reference sentences and may introduce some measurement noise. Therefore, our experiments use method~(b) to compute document-level scores, hereafter referred to as \textbf{ds-BLEU} scores.} 
\begin{figure}[h]
    \begin{minipage}[c]{\linewidth}
    \centering
    \includegraphics[width=0.92\textwidth]{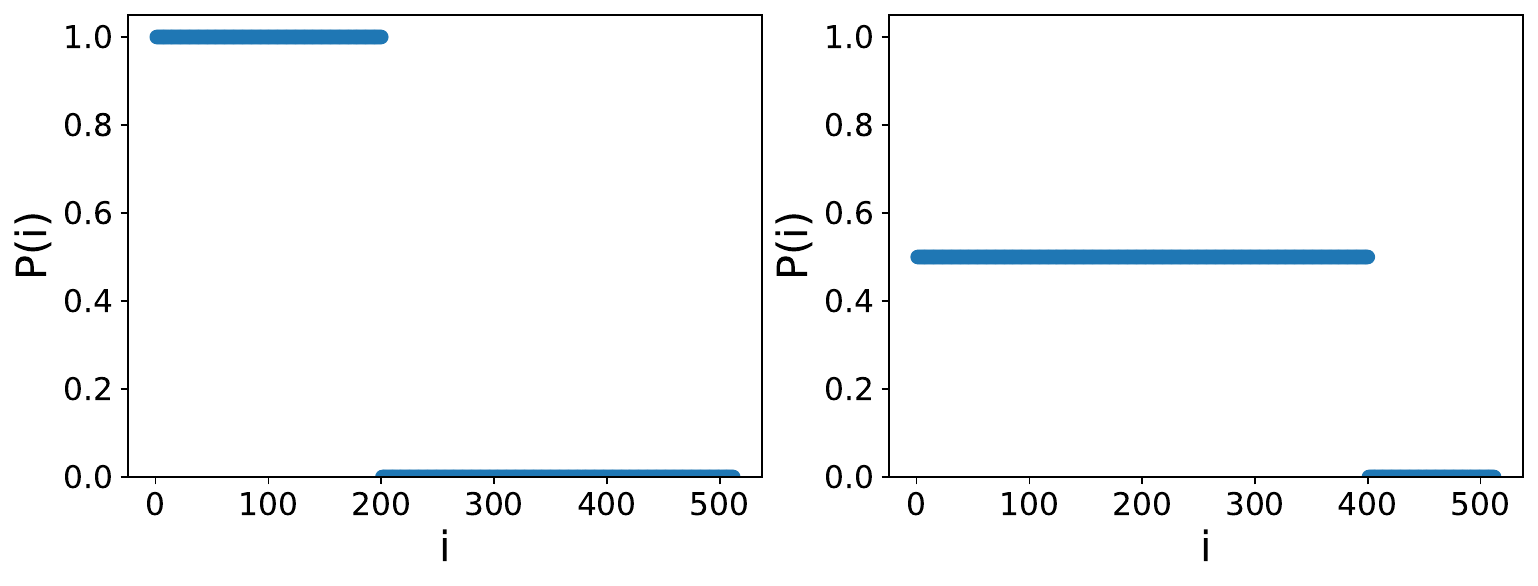}
    \includegraphics[width=0.92\textwidth]{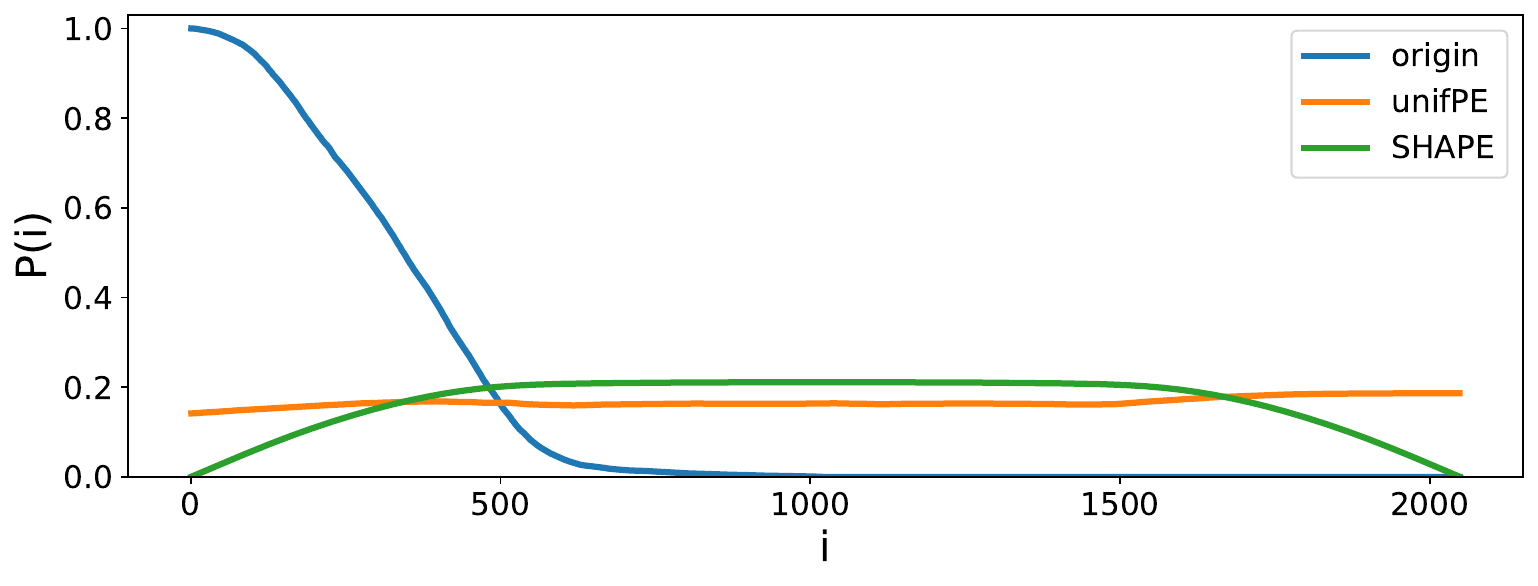}
    \includegraphics[width=0.92\textwidth]{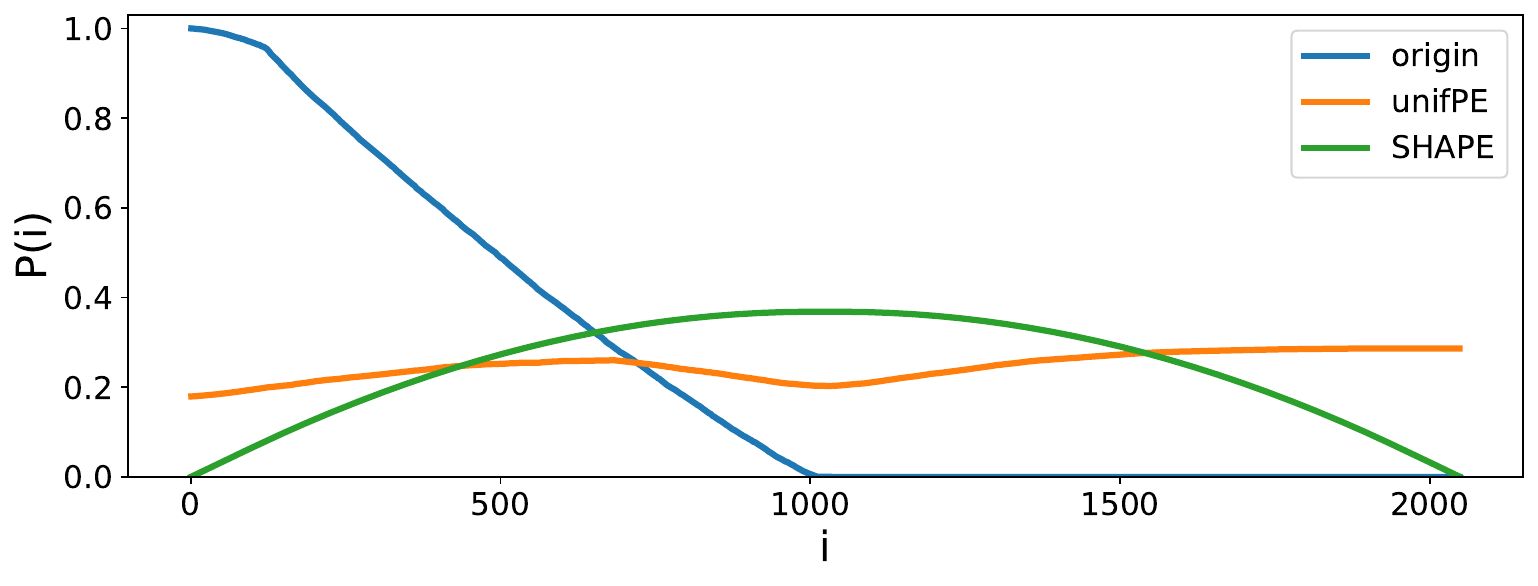}
    \end{minipage}
    \caption{Top: probability of observing training position $i$ ($P(i)$) for a sentence of length $l=200$, with standard training ($k_i=0$, left) and with our uniform sampling scheme (right) for $M = 512$. 
    Middle: \textcolor{blue}{original}, \textcolor{orange}{\posUnif}, and \textcolor{teal}{\model{SHAPE}} $P(i)$ for training set \tedII and $M = 2048$.
    Bottom: \textcolor{blue}{original}, \textcolor{orange}{\posUnif} and \textcolor{teal}{\model{SHAPE}}  $P(i)$ for \tedI and $M = 2048$.
    }
    \label{fig:posunif-prob}
\end{figure}
\subsection{Evaluating Length Issues in MT \label{ssec:methods-length}}
Another recurring methodological caveat with length-related evaluation is related to the way scores are compared. For instance, in \citep[Figure~1]{sun-etal-2022-rethinking} BLEU scores are reported for buckets of sentences of varying lengths in a plot which suggests that performance increases with length (up to a certain extent). Such vizualisations are misleading, as global BLEU scores should only be compared when measured with the same corpus. 

What we propose instead is to compare matching automatic translation scores for a set of inputs $\mathcal{S} = \{s_1 \dots s_T\}$, systematically varying the translation models $M$ in $\{M_1 \dots M_N\}$ and the length of the translation window $W \in \{W_1\dots W_K\}$. For each pair of settings, we can perform a paired t-test for the average score difference and decide whether two configurations $(M_i,W_k)$ and $(M_j,W_l)$, each associating a system and a length, are statistically different, and if so, which of the two is the best.

In our experiments, we consider two ways of presenting $\mathcal{S}$: (a)~at the document level, where each $s_i$ is a document and the evaluation is the ds-BLEU score introduced in Section~\ref{ssec:bleus},\footnote{We use SacreBLEU \citep{post-2018-sacrebleu} with signature: \newline
nrefs:1$|$case:mixed$|$eff:no$|$tok:13a$|$smooth:exp$|$version:2.4.0; the parameter \emph{eff} is set to yes for ds-BLEU.} and (b)~at the sentence level, where each $s_i$ is a sentence and the associated metric is COMET \citep{rei-etal-2020-comet}.\footnote{Using the library \url{https://github.com/Unbabel/COMET} with the default model \texttt{wmt22-comet-da}.} 
\edite{For~(b) we need to realign translation hypotheses with their references. This can be performed with the method of \citet{wicks-post-2022-sentence},\footnote{\citet{junczys-dowmunt-2019-microsoft}'s approach includes a set of tags that constrain input and output to have the same number of sentences, see also \citep{li-etal-2023-ptransformer}.} or with that of \citet{matusov-etal-2005-evaluating},\footnote{\url{https://www-i6.informatik.rwth-aachen.de/web/Software/mwerSegmenter.tar.gz}/.}} which has long been used for evaluating speech translation systems, and which we adopt.\footnote{\edite{The per-sentence COMET scores are averaged at the document level to be associated with document lengths, or at the corpus level to assess global translation quality.}}
Variations in configurations $(M,W)$ are obtained by changing the translation engine and the length of input source texts. In all cases, score comparisons are performed on identical source texts.

\done{Also position: check}
The same technique is also used to measure the impact of the position within a document on translation quality. The question we study is whether quality remains constant across a document, or whether it tends to decrease when sentences are processed at higher position indices. For this, we consider groups of sentences translated at varying starting positions with multiple systems and compare the differences between COMET scores with a paired difference test. 
Details regarding the corpus and window sizes are given in Section~\ref{ssec:exp-data}.

\section{Manipulating the Distribution of PEs}
\label{sec:posunif}
A basic requirement for document-level systems is that they should be trained, or at least fine-tuned, with long text inputs, ideally with complete documents. Using the empirical document length distribution may however not be ideal, as it yields very skewed distributions of PEs where small position indices are over-represented. We discuss two approaches to obtain more balanced distributions.

\subsection{Distribution of PEs \label{ssc:posunif-meta} }
\done{Explain there are 2 ways: changing the pseudo length; shifting PEs} 
A training sequence of length $l$ yields examples for all indices in $\{1, \dots, l \}$. For a complete corpus, 
position index $1$ will be observed for all inputs, while the last index of the longest sequence will likely only be observed once. Training with the ``natural'' distribution of document lengths is therefore likely to overfit to smaller position indices while underfitting to larger ones, hindering the ability to handle long texts or extrapolate to lengths unseen in training \citep{peng2024yarn,zhu2024pose}.

A first way to improve the distribution of token positions seen in training is to increase the representation of long documents in the training data while keeping a good balance with shorter ones \citep{bao-etal-2021-g,sun-etal-2022-rethinking}. This is easy to do in our controlled setting (see Section~\ref{ssec:exp-data}). As our experiments show, 
this significantly improves automatic scores for the context lengths seen during training. 
An alternative, which allows us to better study the effect of PE distributions in training, is to directly manipulate the indices (for a fixed length distribution). 
The \posUnif algorithm, introduced below, is one way to achieve this.

\subsection{Uniform SHAPEs (\posUnif) \label{ssc:algo} }
We assume a training set of texts $s_1 \dots s_N$ of respective lengths $l_1 \dots l_N$, and a maximum model length of $M$, with $\forall i, M > l_i $. Training with text $s_i$ creates training samples for positions $i$ in $[1\!:\!l_i]$. For the whole corpus, positions from $1$ (observed $N$ times) to $l_{max}=\mathrm{max}_{i=1 \dots N}(l_i)$ are observed, with larger indexes being less trained than smaller ones. Positions indices in $[l_{max}\!:\!M]$ are never observed. We wish to make the training PE distribution more even, so that all positions in $[1\!:\!M]$ are equally well-trained, which should also help to extrapolate PEs for indices larger than $l_{max}$.

This can be achieved by shifting the starting index of every $s_i$ by some offset $k_i$, making it possible to train with PEs in $[1+k_i\!:\!l_i+k_i]$. How should $k_i$ be chosen? Randomly choosing $k_i=0$ or $k_i=l_i$ with probability $1/2$ makes the probability of observing any index in $[1\!:\!2l_i]$ equal to $1/2$. This can be generalised to choose $k_i$ with uniform probability $1/m$ among $\{0, l_i, \dots, (m-1)*l_i\}$, with $m= \lfloor M / l_i \rfloor$. However, doing so implies that indices in $[m*l_i:M]$ are never observed. We compensate for this as follows: before sampling $k_i$, we modify the set of possible shifts by adding $r_i = M - m*l_i$ to all values larger than a random index $l \in [1\!:\!m]$. In other words, $k_i$ is sampled from $\{j*l_i+r'_{i,j}, j=0 \dots m-1\}$, with $r'_{ij} = 0$ if $j < l$ and $r_i$ otherwise. Sampling $k_i$ independently for each text $s_i$ in each training batch ensures that all indices are uniformly represented. A formal description of \posUnif is given in Algorithm~\ref{algo:uniform-pos}. Figure~\ref{fig:posunif-prob} illustrates the difference between always starting at position $1$ ($\forall i, k_i=0 $) and using our \posUnif strategy.

This approach is reminiscent of \model{SHAPE} \citep{kiyono-etal-2021-shape}; while \model{SHAPE} chooses
the offset $k_i$ uniformly at random in a fixed interval to simulate relative PEs, \rebuttal{which reduces the frequency of small position indices,} we sample $k_i$ non-uniformly to ensure that all indexes are equally represented in training. 

\section{Experimental Settings \label{sec:exp}}
\subsection{Datasets \label{ssec:exp-data}}
For our experiments, we prepare multiple sets of parallel pseudo-documents based on the EN--FR part of the TEDtalks corpus \citep{cettolo-etal-2012-wit3}. 
\begin{table*}
\begin{floatrow}
    \scalebox{0.82}{
    \small
    
        \begin{tabular}{l|rr|rr|rr}
        \toprule
         & \multicolumn{2}{c|}{\corpus{TED-full}} & \multicolumn{2}{c|}{\tedII} & \multicolumn{2}{c}{\tedI} \\
         & train & dev & train & dev & train & dev \\
         \midrule
        Count & 1831 & 19 & 15625 & 160 & 10582 & 106 \\
        Length & 2915 & 2861 & 341 & 339 & 504 & 512 \\
        \bottomrule
        \end{tabular}
    }
    {%
    }
    \hspace{0.03\textwidth}

    \scalebox{0.82}{
    \small
    \begin{tabular}{l|rrrrrrrrr}
    \toprule
     & sent & 256 & 512 & 768 & 1024 & 1200 & 1600 & 2048 & doc \\
     \midrule
    Count & 5103 & 503 & 261 & 184 & 142 & 123 & 100 & 80 & 52 \\
    Length & 23 & 233 & 450 & 638 & 827 & 955 & 1175 & 1468 & 2259 \\
    \bottomrule
    \end{tabular}
        }
\caption{Left: Statistics of the TED talks training and dev sets. Right: Statistics of the TED talks test sets from IWSLT \testset{tst2014}, \testset{tst2015}, \testset{tst2016} and \testset{tst2017}. `Count' denotes the number of parallel pseudo-documents, `Length' denotes the average length of source (i.e.\ English) pseudo-documents (in \model{NLLB} tokens).
}
\label{tab:data-stat}
\end{floatrow}

\end{table*}

\begin{table}
  \centering
    \scalebox{0.76}{
    \begin{tabular}{lr|rrrr}
    \toprule
    & $l_{max}$ & 2014 & 2015 & 2016 & 2017 \\
    \midrule
    \parbox[t]{3mm}{\multirow{5}{*}{\rotatebox[origin=c]{90}{\model{NLLB}}}} &
    sent & 45.1 \footnotesize{(0.97)} & 43.9 \footnotesize{(0.98)} & 41.7 \footnotesize{(1.00)} & 41.8 \footnotesize{(1.00)} \\
    & 256 & 33.9 \footnotesize{(0.82)} & 35.4 \footnotesize{(0.84)} & 33.3 \footnotesize{(0.86)} & 33.5 \footnotesize{(0.87)} \\
    & 512 & 14.6 \footnotesize{(0.44)} & 16.0 \footnotesize{(0.56)} & 15.2 \footnotesize{(0.52)} & 13.8 \footnotesize{(0.49)} \\
    & 768 & 7.3 \footnotesize{(0.27)} & 7.9 \footnotesize{(0.32)} & 10.0 \footnotesize{(0.46)} & 6.7 \footnotesize{(0.27)} \\
    & 1024 & 8.8 \footnotesize{(0.56)} & 7.4 \footnotesize{(0.51)} & 7.5 \footnotesize{(0.50)} & 6.5 \footnotesize{(0.48)} \\

     \midrule
    \parbox[t]{3mm}{\multirow{8}{*}{\rotatebox[origin=c]{90}{\model{TowerBase}}}} &
    sent & 43.4 \footnotesize{(0.98)} & 42.9 \footnotesize{(0.99)} & 39.7 \footnotesize{(1.00)} & 38.7 \footnotesize{(1.00)} \\
    & 256 & 44.0 \footnotesize{(0.96)} & 42.8 \footnotesize{(0.98)} & 40.9 \footnotesize{(1.00)} & 39.4 \footnotesize{(1.00)} \\
    & 512 & 42.9 \footnotesize{(0.96)} & 39.8 \footnotesize{(0.98)} & 39.9 \footnotesize{(1.00)} & 40.6 \footnotesize{(1.00)} \\
    & 768 & 39.6 \footnotesize{(0.98)} & 39.0 \footnotesize{(0.97)} & 38.1 \footnotesize{(0.99)} & 39.9 \footnotesize{(1.00)} \\
    & 1024 & 38.5 \footnotesize{(0.98)} & 33.1 \footnotesize{(0.99)} & 35.4 \footnotesize{(1.00)} & 35.4 \footnotesize{(0.98)} \\
    & 1200 & 37.4 \footnotesize{(0.92)} & 35.5 \footnotesize{(0.98)} & 36.2 \footnotesize{(1.00)} & 35.6 \footnotesize{(0.98)} \\
    & 1600 & 33.3 \footnotesize{(0.96)} & 34.9 \footnotesize{(0.96)} & 26.7 \footnotesize{(0.94)} & 31.0 \footnotesize{(0.97)} \\
    & 2048 & 24.0 \footnotesize{(0.97)} & 27.7 \footnotesize{(0.95)} & 27.2 \footnotesize{(0.96)} & 23.5 \footnotesize{(0.87)} \\
    \bottomrule
    \end{tabular}
    
    }
  \caption{ds-BLEU scores (and brevity penalty) for \model{NLLB200-distilled-600M} and \model{TowerBase-7b}.}
  
  \label{tab:res-base-ds-bleu}
\end{table}

\begin{table*} 
    \centering
  \setlength{\tabcolsep}{2.9pt}
    \begin{minipage}{0.48\textwidth}
    \scalebox{0.65}{
    \begin{tabular}{rrrrrrrr}
    \toprule
    & \model{NLLB} & FT-U & Unif-U & SHAPE-U & FT-G & Unif-G & SHAPE-G \\
    \midrule
    sent-256 & 9.2 & 0.8 & -2.1 & -2.5 & \pvalue{0.4} & -0.7 & -1.7 \\
    256-512 & 19.1 & - & \pvalue{-0.4} & \pvalue{1.4} & - & - & 2.1 \\
    512-768 & 6.9 & - & -0.6 & - & 5.9 & 2.5 & 5.3 \\
    768-1024 & - & \pvalue{0.5} & - & - & 7.2 & 3.0 & 3.7 \\
    1024-1200 & 2.2 & 3.5 & 1.9 & 4.1 & 4.0 & 3.3 & 4.1 \\
    1200-1600 & - & 6.8 & 6.5 & 5.4 & 5.8 & 5.5 & 4.9 \\
    1600-2048 & 1.9 & 5.2 & 4.5 & 3.1 & 4.2 & 5.7 & 6.0 \\
     \midrule
    sent-256 & 16.7 & 3.5 & 1.7 & 1.3 & 2.7 & 2.1 & 1.9 \\
    256-512 & 20.7 & - & \pvalue{-0.4} & 2.4 & 0.6 & - & 4.2 \\
    512-768 & 5.6 & - & - & - & 11.6 & 7.2 & 8.1 \\
    768-1024 & 5.2 & 2.3 & \pvalue{0.8} & - & 10.4 & 7.6 & 7.8 \\
    1024-1200 & - & 7.4 & 4.3 & 6.9 & 3.8 & 4.7 & 4.6 \\
    1200-1600 & 6.1 & 9.6 & 13.4 & 8.3 & 5.4 & 5.5 & 5.6 \\
    1600-2048 & - & 5.1 & 5.0 & 5.9 & 3.9 & 5.6 & 5.0 \\
    \bottomrule
    \end{tabular}
    }   
    \end{minipage}%
    \hspace{0.02\textwidth} 
    \begin{minipage}{0.48\textwidth}
  \setlength{\tabcolsep}{2.9pt}
    \scalebox{0.65}{
    \begin{tabular}{rrrrrrrr}
    \toprule
     & \model{Tower} & FT-U & Unif-U & SHAPE-U & FT-G & Unif-G & SHAPE-G \\
     \midrule
    sent-256 & - & - & - & - & - & - & - \\
    256-512 & - & \pvalue{0.9} & \pvalue{0.8} & \pvalue{0.8} & 0.6 & 0.6 & 0.5 \\
    512-768 & - & - & - & - & \pvalue{0.6} & - & - \\
    768-1024 & 3.4 & - & \pvalue{1.0} & \pvalue{1.2} & \pvalue{1.7} & 1.2 & 2.1 \\
    1024-1200 & - & - & - & - & - & - & - \\
    1200-1600 & 4.7 & 1.7 & 2.1 & - & \pvalue{1.6} & 2.0 & 1.5 \\
    1600-2048 & 5.9 & 7.5 & 6.5 & 7.3 & 8.1 & 7.8 & 7.3 \\
     \midrule
    sent-256 & 3.9 & 2.3 & 2.4 & 2.3 & 2.3 & 2.3 & 2.2 \\
    256-512 & - & \pvalue{0.4} & - & - & \pvalue{0.2} & \pvalue{0.3} & \pvalue{0.3} \\
    512-768 & - & - & - & \pvalue{0.5} & 0.3 & - & - \\
    768-1024 & 2.9 & \pvalue{1.0} & 0.5 & - & \pvalue{1.1} & 0.8 & \pvalue{1.2} \\
    1024-1200 & - & - & \pvalue{1.0} & \pvalue{0.9} & - & - & - \\
    1200-1600 & 6.2 & 1.7 & \pvalue{1.8} & - & - & 1.9 & 1.8 \\
    1600-2048 & 8.7 & 10.0 & 8.9 & 9.1 & 11.0 & 10.2 & 9.2 \\
    \bottomrule
    \end{tabular}
    }
    \caption{Average differences evaluated on \textbf{ds-BLEU} (top) of full TED talks and on $100\times$\textbf{COMET} (bottom) of realigned parallel sentences, \rebuttal{between translations in increasing context size,} for \model{NLLB} (left) and \model{TowerBase} (right) models. U and G respectively denote \tedI and \tedII. \rebuttal{A positive value means that shorter segments result in higher scores than longer ones. Text in \pvalue{olive} for p-values $> 0.01$. - for p-values $> 0.05$.  } }
    \label{tab:ttest-by-block-all}
    \end{minipage}
\end{table*}

\paragraph{Training and validation sets} Our training set consists of pseudo-documents from both the training and validation splits of \corpus{IWSLT-2016}.\footnote{\url{https://wit3.fbk.eu/2016-01}} Our goal is to simulate real corpora of parallel documents with source documents shorter than a certain length $l_{max}$ -- using $l_{max}=1024$. We split all document pairs whose source side is longer than 1024 tokens into fragments.\footnote{All statistics counted in tokens use the tokeniser of \model{NLLB} \citep{costa-etal-2024-nllb}.}
For each document pair, we iterate the following procedure: (1)~sample a maximum pseudo-document length $l'_{i}$ following the same Gaussian-like length distribution as the full TED talks with $l'_{i} < l_{max}$, (2)~concatenate consecutive sentence pairs up to $l'_{i}$ to form a training pseudo-document $s_i$.
The resulting distribution of document lengths is displayed in Figure~\ref{fig:data-distribution} \rebuttal{in Appendix~\ref{ann:data}}.
The development set is built similarly, using document pairs from IWSLT \testset{tst2010} and \testset{tst2011}. We denote these training datasets as \tedII (G for Gaussian).
As discussed in Section~\ref{sec:posunif}, we consider another dataset generation strategy, which produces a more balanced length distribution, for which we do as above but 
we sample uniformly: $l'_{i} \sim U(128, l_{max})$.\footnote{Short pseudo-documents continue to be slightly over-represented, because the last pseudo-document in any given talk is often strictly shorter than the desired length $l'_i$.} Fine-tuning with the resulting \tedI corpus allows us to
contrast two distributions with differences in document length.
\paragraph{Test sets} To evaluate MT systems for their ability to handle documents of varying sizes and extrapolate beyond the training samples, we build a series of test sets of increasing document lengths. For each document in IWSLT \testset{tst2014}, \testset{tst2015}, \testset{tst2016} and \testset{tst2017}, we accumulate consecutive sentence pairs into parallel pseudo-documents such that all resulting source texts have a length close to $l_{max}$, with $l_{max} \in \{256, 512, 1024, 1200, 1600, 2048\}$.\footnote{At the end of each talk, we concatenate the last parallel sentences into the last pseudo-document if they are shorter than $50$ to avoid exceedingly short parallel sequences.} Contrarily to training sets, test sets are homogeneous in length. Statistics are in Table~\ref{tab:data-stat} with more details \rebuttal{in Appendix~\ref{ann:data}}. Evaluation is always performed with complete original talks, after concatenating and aligning all the corresponding parts. 
\subsection{Models \label{ssec:exp-model}}
We used the \posUnif algorithm to fine-tune two pre-trained MT systems \edite{that were not trained with TED talks.}
As \posUnif is designed for APEs, we considered
\model{NLLB200-distilled-600M}\footnote{\url{https://huggingface.co/facebook/nllb-200-distilled-600M}} or \model{NLLB} for short \citep{costa-etal-2024-nllb} as a representative encoder-decoder model based on APEs. \model{NLLB} is a 12-layer encoder-decoder multilingual MT model pre-trained on 200~languages. 
We used the HuggingFace implementation, which relies on sinusoidal APEs \citep{vaswani-etal-2017-attention}.
\rebuttal{We also perform fine-tuning with \model{SHAPE} for comparison.}
\rebuttal{We refer to the specific MT systems with respect to their fine-tuning method (FT, \posUnif or \model{SHAPE}), backbone model (e.g.\ \model{NLLB}) and training corpus (U for \tedI or G for \tedII. More precisely,}
we denote MT systems trained on \tedI (resp. \tedII) as \FTnllbI (resp.\ \FTnllbII), \UNIFnllbI (resp.\ \UNIFnllbII) when fine-tuning with \posUnif, \rebuttal{and \SHAPEnllbI (resp.\ \SHAPEnllbII)}.

We also experiment with an LLM-based architecture, \model{TowerBase-7B}\footnote{\url{https://huggingface.co/Unbabel/TowerBase-7B-v0.1}} \citep{alves2024tower} (\model{TowerBase} for short), derived from Llama2 \citep{touvron2023llama} using translation-related tasks. \model{TowerBase} uses RoPE \citep{SU2024127063} to encode RPEs. As mentioned by \citet{peng2024yarn}, they nonetheless encode some form of APE signal in some dimensions, and may therefore be also mildly impacted by the PE training distribution.
\rebuttal{We refer to the models based on TowerBase as \FTtowerI (resp.\ \FTtowerII), \UNIFtowerI (resp.\ \UNIFtowerII) and \SHAPEtowerI (resp.\ \SHAPEtowerII) the model fine-tuned on \tedI (resp.\ \tedII) with original PEs, \posUnif or \model{SHAPE}.} 

Both backbone models were pretrained with large amounts of EN--FR data; we focus exclusively on the EN into FR direction. \done{Check direction}
%
Details on fine-tuning and decoding parameters can be found in Appendix~\ref{ann:experiments}.
\section{Results and Analyses \label{sec:res}}
\done{put all results about ratio7 in an archiv }
\subsection{Length Issues \label{ssec:res-len}}
We report the \edite{ds-BLEU} (Table~\ref{tab:res-base-ds-bleu}) and COMET \edite{(Appendix, Table~\ref{tab:res-base-comet})} scores
of the pretrained models \model{NLLB} and \model{TowerBase} for multiple test sets, \edite{varying the average input segment lengths from one sentence to the maximum input length used in training}.\footnote{As explained in Section~\ref{ssec:bleus}, these COMET scores require the realignment of target sentences with the reference.}\done{COMET scores in appendix ?}
For \model{NLLB}, we observe a drop of around $10$ \edite{ds-BLEU} points and about $0.2$ COMET points when translating test sets of $l_{max}=256$ instead of isolated sentences. Scores and their associated brevity penalties (BPs) only get worse with larger context lengths. For \model{TowerBase}, the decrease in BLEU is more progressive, with a sharp decline for all test sets for $l_{max} > 1024$. The related COMET scores plummet immediately with a context size of $256$. Even though \model{TowerBase} is based on Llama2, which accepts inputs up to~$4096$ tokens, the continued pretraining that was used mostly uses isolated sentences, which introduces an inductive bias affecting its ability to translate long texts.

As expected, document-level fine-tuning (DLFT)
has a strong positive impact (see Appendix, Table~\ref{tab:ttest-compare-system-base-ft}). However, the length issues remain.

\begin{table*}
  \centering
  \setlength{\tabcolsep}{6pt}
    \scalebox{0.7}{
    \begin{tabular}{r|rr|rr|rrr}
    \toprule
     & \multicolumn{2}{c|}{\tedI} & \multicolumn{2}{c|}{\tedII} & FT & Unif & SHAPE \\
     & \ FT vs Unif \ & \ FT vs SHAPE \ & \ FT vs Unif \ & FT vs SHAPE & U vs G & U vs G & U vs G \\
     
     \midrule
    sent & 3.3 \footnotesize{(0.00)} & 4.0 \footnotesize{(0.00)} & 1.2 \footnotesize{(0.00)} & 2.4 \footnotesize{(0.00)} & - & -2.1 \footnotesize{(0.00)} & -1.6 \footnotesize{(0.00)} \\
    256 & - & 0.7 \footnotesize{(0.00)} & - & - & - & -0.6 \footnotesize{(0.01)} & -0.7 \footnotesize{(0.00)} \\
    512 & -0.5 \footnotesize{(0.01)} & 1.7 \footnotesize{(0.01)} & - & 2.3 \footnotesize{(0.00)} & -0.7 \footnotesize{(0.00)} & \textbf{-0.4 \footnotesize{(0.01)}} & - \\
    768 & -0.8 \footnotesize{(0.00)} & 2.7 \footnotesize{(0.00)} & -3.7 \footnotesize{(0.00)} & - & 5.5 \footnotesize{(0.00)} & 2.6 \footnotesize{(0.00)} & 4.5 \footnotesize{(0.00)} \\
    1024 & - & 1.6 \footnotesize{(0.00)} & -7.8 \footnotesize{(0.00)} & -1.8 \footnotesize{(0.04)} & 12.2 \footnotesize{(0.00)} & 5.1 \footnotesize{(0.00)} & 8.8 \footnotesize{(0.00)} \\
    1200 & -2.3 \footnotesize{(0.00)} & 2.3 \footnotesize{(0.02)} & -8.5 \footnotesize{(0.00)} & - & 12.7 \footnotesize{(0.00)} & 6.5 \footnotesize{(0.00)} & 8.8 \footnotesize{(0.00)} \\
    1600 & \textbf{-2.6 \footnotesize{(0.01)}} & - & -8.8 \footnotesize{(0.00)} & \textbf{-2.6 \footnotesize{(0.01)}} & 11.8 \footnotesize{(0.00)} & 5.6 \footnotesize{(0.00)} & 8.3 \footnotesize{(0.00)} \\
    2048 & \textbf{-3.3 \footnotesize{(0.00)}} & - & -7.3 \footnotesize{(0.00)} & - & 10.7 \footnotesize{(0.00)} & 6.8 \footnotesize{(0.00)} & 11.3 \footnotesize{(0.00)} \\

     \midrule
    sent & 1.9 \footnotesize{(0.00)} & 2.9 \footnotesize{(0.00)} & 0.9 \footnotesize{(0.00)} & 1.4 \footnotesize{(0.00)} & - & -0.9 \footnotesize{(0.00)} & -1.4 \footnotesize{(0.00)} \\
    256 & - & 0.8 \footnotesize{(0.00)} & \textbf{0.4 \footnotesize{(0.00)}} & \textbf{0.7 \footnotesize{(0.00)}} & \textbf{-0.8 \footnotesize{(0.00)}} & -0.5 \footnotesize{(0.01)} & -0.9 \footnotesize{(0.00)} \\
    512 & -0.6 \footnotesize{(0.02)} & 2.9 \footnotesize{(0.00)} & - & 4.3 \footnotesize{(0.00)} & -0.4 \footnotesize{(0.03)} & - & - \\
    768 & -0.7 \footnotesize{(0.00)} & 4.7 \footnotesize{(0.00)} & -4.7 \footnotesize{(0.00)} & - & 11.2 \footnotesize{(0.00)} & 7.2 \footnotesize{(0.00)} & 7.3 \footnotesize{(0.00)} \\
    1024 & \textbf{-2.2 \footnotesize{(0.00)}} & 3.3 \footnotesize{(0.00)} & -7.5 \footnotesize{(0.00)} & -1.8 \footnotesize{(0.02)} & 19.3 \footnotesize{(0.00)} & 14.0 \footnotesize{(0.00)} & 14.2 \footnotesize{(0.00)} \\
    1200 & -5.3 \footnotesize{(0.00)} & 2.7 \footnotesize{(0.02)} & -6.5 \footnotesize{(0.00)} & - & 15.7 \footnotesize{(0.00)} & 14.5 \footnotesize{(0.00)} & 11.9 \footnotesize{(0.00)} \\
    1600 & - & - & -6.4 \footnotesize{(0.00)} & - & 11.5 \footnotesize{(0.00)} & 6.6 \footnotesize{(0.00)} & 9.2 \footnotesize{(0.00)} \\
    2048 & - & \textbf{2.1 \footnotesize{(0.04)}} & -4.7 \footnotesize{(0.00)} & - & 10.4 \footnotesize{(0.00)} & 7.2 \footnotesize{(0.00)} & 8.3 \footnotesize{(0.00)} \\
    \bottomrule
    \end{tabular}
    }
    
    \caption{Average difference (and p-values) in \textbf{ds-BLEU} (top) evaluated on full TED talks and $100\times$\textbf{COMET} (bottom) evaluated on realigned sentences for \model{NLLB}. \rebuttal{
    Left and middle: paired comparison between fine-tuning with the original PEs (FT), \posUnif (Unif) and \model{SHAPE} on \tedI and \tedII respectively. 
    Right: differences between fine-tuning on \tedI (U) and \tedII(G).
    - for p-values $> 0.05$. Bold values when the two metrics disagree on significativity.}
    } 
  \label{tab:ttest-compare-system-nllb}
\end{table*}

\paragraph{Length Bias}
We performed paired comparisons for the translation \edite{of our test sets} with increasing text lengths for each MT system \edite{as presented in Section~\ref{ssec:methods-length}}.
Results are given in Table~\ref{tab:ttest-by-block-all},
where a positive difference (e.g.\ $9.2$ for \model{Nllb} in line ``sent-256'') means that \edite{the translation of} shorter segments (here: sentences) yield better scores than \edite{that of} longer ones ($256$ tokens). 
\edite{Scores in the same column are comparable.}
Except for a handful of configurations, translating longer texts is never better than translating short ones. We conclude that in our experimental settings, the disadvantages associated with long inputs (Section~\ref{ssec:challenge}) overwhelm the benefits of a complete context. These length issues result in large score degradations and are not easily fixed by simple manipulation of PEs.
%
%
We also observe that results obtained with COMET and ds-BLEU sometimes disagree. These cases are rare, though, suggesting that our results are robust. 

\paragraph{Document-level Tuning with \posUnif } 
Again using the paired comparison methodology, we compare the performance of DLFT \rebuttal{with original PEs, \posUnif and \model{SHAPE}}.
As shown in the left \rebuttal{and middle} parts of Table~\ref{tab:ttest-compare-system-nllb}, fine-tuning using \posUnif leads to steady improvements in translation scores for all test lengths, especially for systems fine-tuned with the unbalanced corpus (\tedII). The only exception is for sentence-level translations, which remain marginally better using standard DLFT than with \posUnif. \rebuttal{In contrast, \model{SHAPE} only improves DLFT performance on the \tedII corpus and for translation windows greater than $1024$ tokens, due to the under-representation of small position indices during training, as shown in Figure~\ref{fig:posunif-prob}.} As \Cref{tab:ttest-compare-system-base-ft,tab:res-nllb-ted-distill,tab:res-nllb-ted-unif,tab:res-tower-ted-distill,tab:res-tower-ted-unif} show, these improvements remain moderate, and the length issues continue to strongly impact translation scores, especially for test documents of $1024$ tokens or more. 
\begin{table}
  \centering
    \scalebox{0.69}{
    \setlength{\tabcolsep}{3pt}
    \begin{tabular}{lrrrrrrr}
    \toprule
    & \model{NLLB} & FT-U & Unif-U & SHAPE-U & FT-G & Unif-G & SHAPE-G \\
     \midrule
    $p_0$-$p_1$  & 12.7 & - & - & 4.0 & 1.6 & 3.0 & 5.0 \\
    $p_1$-$p_2$ & 7.2 & - & - & \pvalue{-2.0} & 1.9 & 2.4 & - \\
    $p_2$-$p_3$ & 2.9 & 1.0 & -1.3 & - & 2.4 & - & \pvalue{-2.1} \\
    $p_3$-$p_4$ & 7.2 & 5.5 & 4.6 & 7.3 & 26.1 & 10.7 & 13.9 \\
    $p_4$-$p_5$ & - & 3.9 & - & - & 8.3 & 4.7 & 4.3 \\
    $p_5$-$p_6$  & 3.3 & 31.6 & 27.1 & 19.5 & 6.1 & 15.4 & 15.3 \\
    \bottomrule
    \end{tabular}
    }  

    \medskip
    \setlength{\tabcolsep}{3pt}
    \scalebox{0.69}{
    
    \begin{tabular}{lrrrrrrrrr}
    \toprule
     & \model{Tower} & FT-U & Unif-U & SHAPE-U & FT-G & Unif-G & SHAPE-G \\
     \midrule
    $p_0$-$p_1$ & \pvalue{1.2} & - & - & - & - & - & - \\
    $p_1$-$p_2$ & - & - & - & - & \pvalue{0.6} & \pvalue{0.6} & - \\
    $p_2$-$p_3$ & - & - & - & - & - & - & - \\
    $p_3$-$p_4$ & 4.9 & 1.1 & \pvalue{0.6} & 1.4 & 1.1 & 1.3 & 1.1 \\
    $p_4$-$p_5$ & \pvalue{1.7} & 1.4 & 2.1 & 1.7 & 2.0 & 1.8 & 2.2 \\
    $p_5$-$p_6$ & 26.3 & 24.5 & 26.0 & 25.2 & 27.1 & 26.4 & 27.0 \\
    \bottomrule
    \end{tabular}
    } 

    \caption{Average difference of $100\times$COMET-score evaluated on $794$ sentence pairs, translated at different positions \rebuttal{(e.g.\ $p_0$ and $p_1$ with $p_0<p_1$)} by \model{NLLB}-based systems (top) and \model{TowerBase}-based systems (bottom). \rebuttal{\pvalue{Olive} text for p-values~$> 0.01$. - for p-values~$> 0.05$. }  
    }
    %
  \label{tab:ttest-position-nllb-tower}
\end{table}
For \model{TowerBase}, \posUnif does not yield any significant difference with standard DLFT, \rebuttal{and \model{SHAPE} occasionally delivers slight improvements} (see Appendix, Table~\ref{tab:ttest-compare-system-tower}), likely because this model relies on RPEs. From these comparisons, we conclude that \posUnif partly resolves length issues for \model{nllb}, but hardly changes the situation for \model{TowerBase}.
\paragraph{Impact of Data Distribution }
In the right part of Table~\ref{tab:ttest-compare-system-nllb}, we evaluate the impact of the length distribution during fine-tuning \edite{for \model{NLLB}}: the balanced distribution (\tedI) \edite{slightly but} consistently underperforms the use of \tedII for short documents (fewer than $512$ tokens), a trend that is reversed for longer documents \edite{with strong improvement} (over $768$ tokens). Manipulating the distribution of PEs with \posUnif reduces the gap between the two fine-tuning corpora and makes the model more robust to document lengths rarely observed (or even unobserved) during fine-tuning. 
This analysis again reveals small differences between using ds-BLEU and COMET scores: in nine cases out of $56$ comparisons (marked in bold), one metric detects a difference that is non-significant for the other.

\subsection{Position Bias \label{ssec:res-position}}
To investigate potential translation issues related to large position indices, we collected the $794$ sentences \edite{that come from the final part of long talks and }for which varying the window length also varied the position index.
For each of them, we have seven translations, corresponding to positions $\{p_0^j, \dots, p_6^j\}, j \in \{1, \dots, 794\}$. The average values for $\{p_0^j, \dots, p_6^j\}$  are $\{p_0, \dots, p_6\} = [66, 173, 262, 335, 585, 779, 1477]$.
For this subset of sentences, we performed a paired t-test to compare the impact of the position on the translation score (using COMET as the only metric).
We observe in Table~\ref{tab:ttest-position-nllb-tower} that in almost all comparisons but three, a small position index is preferable to a larger one. This suggests that one of the main challenges faced by \emph{Doc2Doc} with large context lengths is to control the quality degradation \edite{for} the final parts of the input text. Here again, 
\posUnif slightly mitigates the problem for \model{NLLB} models \rebuttal{compared with original PEs and \model{SHAPE}}, but no such improvement is observed for \model{TowerBase}.\done{Why is tower different?}

\begin{table}
  \centering
    \scalebox{0.7}{

    \begin{tabular}{lrrrrlll}
    \toprule
     & 256 & 512 & 768 & 1024 & 1200 & 1600 & 2048 \\
     \midrule
    \model{NLLB} & 0.04 & 0.35 & 0.49 & 0.66 & 0.64 & 0.74 & 0.81 \\
    FT-U & 0.01 & 0.03 & 0.08 & 0.09 & 0.13 & 0.26 & 0.44 \\
    Unif-U & 0.01 & 0.03 & 0.07 & 0.10 & 0.20 & 0.34 & 0.32 \\
    SHAPE-U & 0.03 & 0.08 & 0.11 & 0.20 & 0.36 & 0.39 & 0.46 \\
    FT-G & 0.01 & 0.03 & 0.11 & 0.31 & 0.40 & 0.57 & 0.69 \\
    Unif-G & 0.01 & 0.04 & 0.16 & 0.20 & 0.27 & 0.25 & 0.36 \\
    SHAPE-G & 0.02 & 0.05 & 0.12 & 0.17 & 0.21 & 0.24 & 0.28 \\
    \bottomrule
    \end{tabular}
    }  
    \medskip
    
    \scalebox{0.7}{
    \begin{tabular}{lrrrrrrr}
    \toprule
     & 256 & 512 & 768 & 1024 & 1200 & 1600 & 2048 \\
     \midrule
    \model{Tower} & 0.01 & 0.05 & 0.13 & 0.30 & 0.29 & 0.45 & 0.64 \\
    FT-U & 0.01 & 0.05 & 0.10 & 0.15 & 0.13 & 0.23 & 0.59 \\
    Unif-U & 0.02 & 0.05 & 0.09 & 0.15 & 0.16 & 0.28 & 0.61 \\    
    SHAPE-U & 0.02 & 0.05 & 0.08 & 0.15 & 0.17 & 0.21 & 0.59 \\
    FT-G & 0.02 & 0.05 & 0.10 & 0.15 & 0.19 & 0.26 & 0.62 \\
    Unif-G & 0.02 & 0.07 & 0.10 & 0.16 & 0.20 & 0.28 & 0.64 \\
    SHAPE-G & 0.02 & 0.05 & 0.07 & 0.17 & 0.17 & 0.27 & 0.62 \\
    \bottomrule
    \end{tabular}

    }  
    \caption{Percentage of pseudo-documents among IWSLT \testset{tst2014-2017} in which $10$-gram repetition is detected in the translation given by \model{NLLB}-based (top) and \model{TowerBase}-based models (bottom).}
  \label{tab:stat-repeat}
\end{table}
\subsection{Repeated $n$-grams in Translation\label{ssec:res-repeat}}
One obvious problem with holistic translations produced by \model{Nllb} is the generation of outputs that are too short. A closer look at translation outputs also reveals that outputs contain many instances of repeated texts, usually occurring in the final part of the translation. 
To quantify this problem, we compute the percentage of translations of pseudo-documents in which the repetition of a long $n$-gram (with $n \ge 10$) is detected. Detailed results are given in Table~\ref{tab:stat-repeat}. For all systems and fine-tuning strategies, the percentage of repetitions increases with the length, a problem that seems (for large text lengths) slightly more severe for \model{TowerBase}\edite{, which has a much better BP,} than \model{Nllb}. 
\section{Conclusion \label{sec:conclusion}}\zpTodo{More criticisms of BLEU}
In this work, we have studied the ability of current MT architectures to handle long input texts, ideally entire documents, and to translate them holistically. Our analyses are based on systematic comparisons of translation outputs computed with varying input lengths, which are then evaluated with two automatic metrics. They consistently show that, even when the test document lengths match that of the training set and remain within the model limits, the translation scores tend to decrease with the source length, a degradation that mostly impacts sentences occurring far from the beginning of the document. We also show that manipulating the training distribution of lengths or PEs has a positive effect for APE-based models, which vanishes in \edite{RoPE-based} models like \model{TowerBase}.\done{Make sure RPE is defined} These results finally confirm the robustness of sentence-level baselines. They hint at the need to improve existing models to truly benefit from the potential of document-level MT, for instance by constraining the attention mechanism to simulate a form of sentence alignment, by improving the memorization capacities of existing architectures, or by ensuring that the generation algorithm does not eventually get trapped in repetition loops. These are some of the directions we wish to explore in future work.\done{Perhaps: more citations for this}
\zpTodo{Future: the errors beyond repetitions and missing material}
\done{there should not be TEDtalks in NLLB training set, as the NLLB papers used IWSLT TED talks as benchmark to show the model performance, where is the only mention os TED talks.}
\section{Limitations \label{sec:limitations}}
The empirical observations reported in this paper are based on one single language direction,\done{check that this is clear} and one domain (TEDtalks). This experimental design is motivated by (a) the fact that French-English is considered an easy pair for MT, with large sets of parallel training data; (b) TEDtalks data are a standard benchmark for document-level MT, and crucially contain very long parallel documents, allowing us to implement our evaluation methodology on a large range of length values. Furthermore, these datasets are not included in the training data of our models. We contend that the length issues observed in these favorable conditions for two representative systems would only be worse for more difficult or less-resourced language pairs.

\section{Ethics Statement \label{sec:ethics}}
This study has been performed with standard benchmarks and open-weight models. We do not see any ethical problems with this work.

\section{Carbon Impact Statement \label{sec:co2} }
\edite{The experiments were conducted on a private infrastructure using a single A100 SXM4 GPU, with a carbon efficiency of $0.432$ kgCO$_2$eq/kWh. The average time required for fine-tuning and checkpoint selection was $14.21$ hours for the six \model{NLLB} models, and $5.6$ hours for the \model{Towerbase} models.
The average emissions are estimated to be $2.45$ kgCO$_2$eq for \model{NLLB}-based models and $0.97$ kgCO$_2$eq for models derived from \model{TowerBase}, with no offset applied.
These estimations were based on the Machine Learning Impact calculator}\footnote{\url{https://mlco2.github.io/impact\#compute}} \citep{lacoste2019quantifying}. 

\section*{Acknowledgments}
\rebuttal{This work was supported by the French national agency ANR as part of the MaTOS project.\footnote{\url{http://anr-matos.fr/}} Rachel Bawden was also partly funded by her chair position in the PRAIRIE institute funded by ANR as part of the ``Investissements d’avenir'' programme under reference ANR19-P3IA-0001.
The authors are grateful to the anonymous reviewers for their insightful comments and suggestions and to Paul Lerner for his review and feedback on a preliminary draft of this work.}

\bibliography{matos}

\appendix
\section{Appendix}
\label{sec:appendix}

\subsection{The \posUnif Algorithm \label{ann:algo}}

The \posUnif algorithm briefly described in Section~\ref{ssc:algo} is formalised in Algorithm~\ref{algo:uniform-pos}.
\begin{algorithm}
    \DontPrintSemicolon
    \small
    \KwData{$l_i$: The input length}
    \KwData{$M$: The target max context length}
    \KwData{$\operatorname{List_{p_k}}$: the distribution of $p_k$ for each offset $k$ in $[0, M-l_i]$}
    
    $\operatorname{List_{p_k}} \longleftarrow \text{Initialized to $0$ for each element}$

    $m \longleftarrow \lfloor M / l_i \rfloor\text{ nb.\ of possible non-zero } p_k $

    $R_n \longleftarrow \text{the remainder of $M$ divided by }l_i$

    $p_0 \longleftarrow \frac{1}{m} \text{the probability of each non-zero $p_k$ }$

    \eIf{$M < 2l_i$}{
        $\operatorname{List_{p_k}} \longleftarrow  p(k'=0) = 1 \text{ i.e.\ } p_{k=0} = 1  $
    }{
    
        $k^* \longleftarrow \text{a random integer in $[0, m)$}$

        \For{$k \in [0, M-l_i]$}{
            \If{$k\%l_i==0$ \texttt{ and } $k < k^*$}{
               $\operatorname{List_{p_k}} \longleftarrow  p(k'=k) = p_0$
            }
            
            \If{$(k - R_n)\%l_i == 0$ \texttt{ and } $k^* < k \leq M-l_i$}{
                $\operatorname{List_{p_k}} \longleftarrow  p(k'=k) = p_0$
            }
            
        }
    }
    \Return $\operatorname{List_{p_k}}$ 
    \newline
    \caption{\posUnif: the pseudo-uniform position indices mapping algorithm.}
    \label{algo:uniform-pos}  
\end{algorithm}

\subsection{A Call for Correctly Using BLEU Scores \label{ann:bleu}}
\begin{figure}[h]
    \begin{minipage}[c]{\linewidth}
    \centering
    \includegraphics[width=0.7\textwidth]{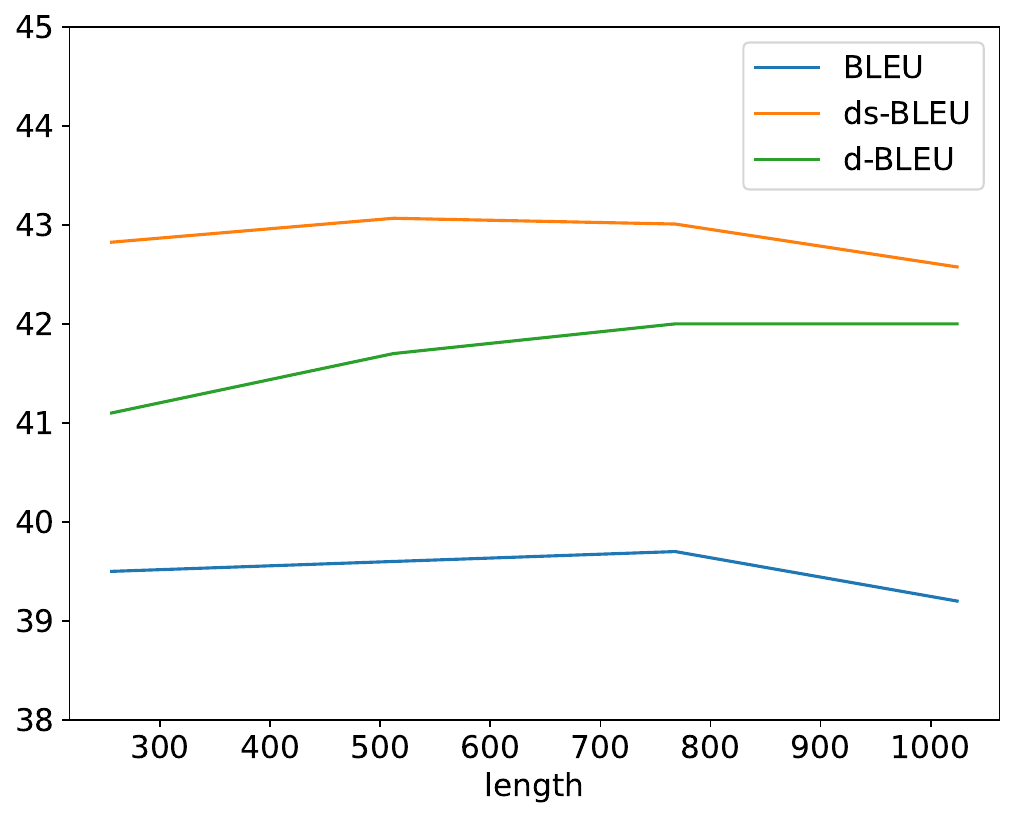}
    \includegraphics[width=0.7\textwidth]{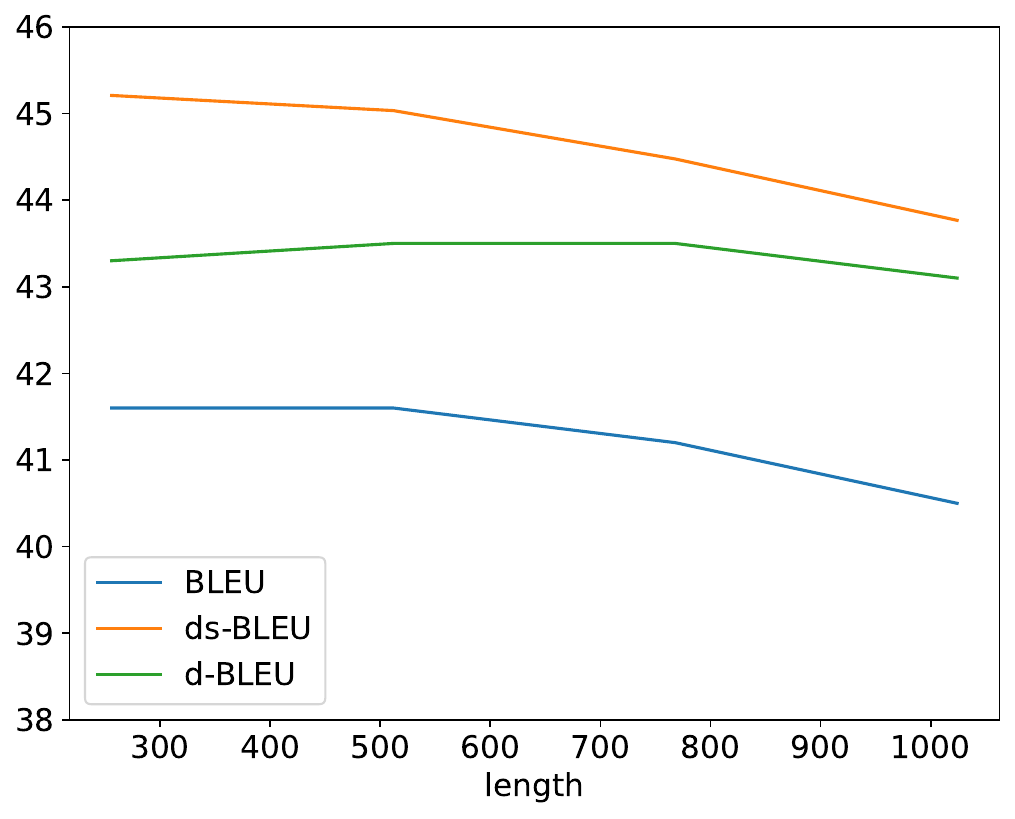}
    \end{minipage}
    \caption{BLEU, ds-BLEU and d-BLEU scores for IWSLT \testset{tst2015}, translating and evaluating \emph{pseudo-documents} of increasing lengths $[256, 512, 768, 1204]$, using \FTnllbI (top) and \UNIFtowerI (bottom). Note that d-BLEU is computed for pseudo-documents while ds-BLEU is computed for concatenated full talks.}
    \label{fig:bleu}
\end{figure}

\edite{As illustrated in Figure~\ref{fig:bleu}, d-BLEU and ds-BLEU are always larger than BLEU. When BLEU decreases due to the degradation of translation quality, {d-BLEU} remains stable because of the higher probability to find $n$-gram matches in longer sequences. In contrast, ds-BLEU consistently decreases when BLEU diminishes, as it applies a macro-average to compute the corpus-level score, which is more sensitive to the translation quality of each document than d-BLEU. Therefore, d-BLEU, ds-BLEU and BLEU are not comparable and d-BLEU is not suitable for analysing length issues in document-level evaluation of MT.}

\begin{figure}[h]
    \begin{minipage}[c]{\linewidth}
    \centering
    \includegraphics[width=0.7\textwidth]{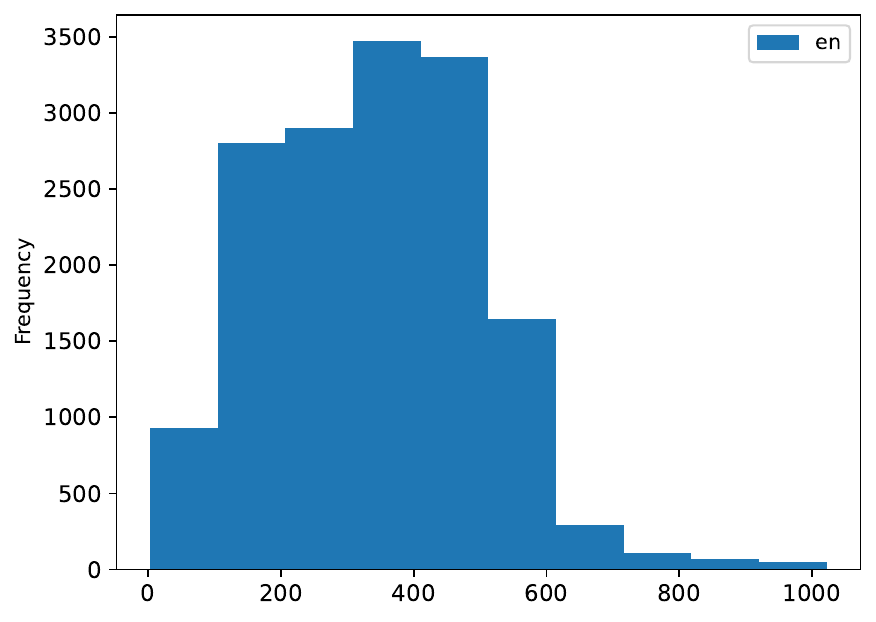}
    \includegraphics[width=0.7\textwidth]{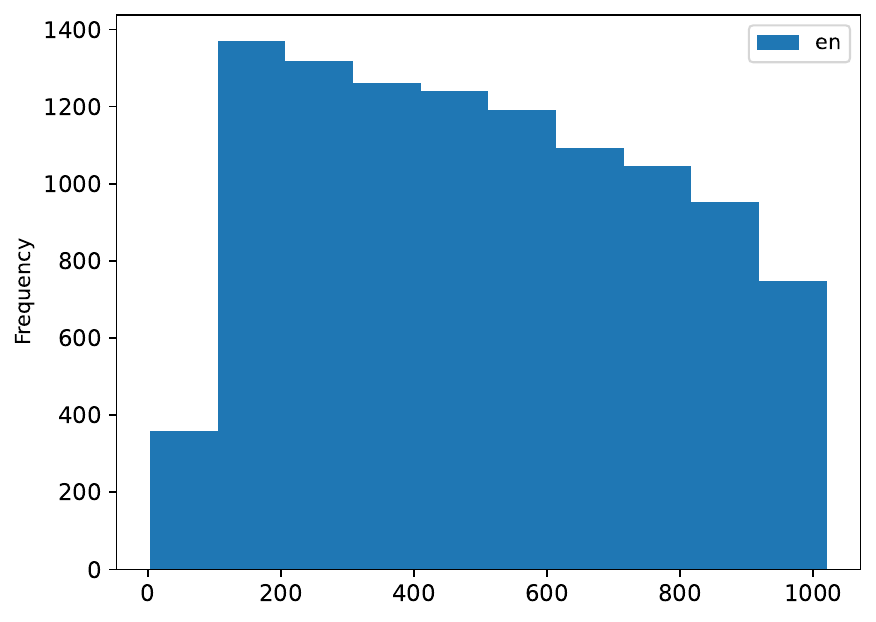}
    \end{minipage}
    \caption{Source document length distribution in the training set of \tedII (top) and \tedI (bottom).  }
    \label{fig:data-distribution}
\end{figure}

\subsection{Data Statistics \edite{and Other Details} \label{ann:data}}
Full data statistics are given in Tables~\ref{tab:data-train-full} and~\ref{tab:data-testsets}.
\edite{All the full TED talks in our corpora start with the title, then the description and the talk before being split into pseudo-documents. \texttt{<description>} and \texttt{<title>} tags are removed.
}
\edite{When preparing our training and validation sets \tedI and \tedII (see Section~\ref{ssec:exp-data}), if concatenating the last sentence pair $(x_n, y_n)$ into the current pseudo-document pair exceeds $l_{max}$, $(x_n, y_n)$ will yield a single parallel sequence, to respect the maximum length $l_{max}$. 
The length distribution is illustrated in Figure~\ref{fig:data-distribution}.}

\begin{table}
  \centering
    \scalebox{0.7}{
    \begin{tabular}{lr|rrrr}
    \toprule
     & & 2014 & 2015 & 2016 & 2017 \\
     \midrule
    \parbox[t]{3mm}{\multirow{5}{*}{\rotatebox[origin=c]{90}{\model{NLLB}}}} &
    sent & 84 & 85 & 85 & 84 \\
    & 256 & 68 & 69 & 68 & 66 \\
    & 512 & 49 & 47 & 47 & 46 \\
    & 768 & 43 & 42 & 40 & 41 \\
    & 1024 & 36 & 37 & 36 & 36 \\

     \midrule
    \parbox[t]{3mm}{\multirow{8}{*}{\rotatebox[origin=c]{90}{\model{TowerBase}}}} &
    sent & 84 & 85 & 85 & 85 \\
    & 256 & 80 & 81 & 82 & 80 \\
    & 512 & 79 & 80 & 82 & 80 \\
    & 768 & 78 & 78 & 80 & 80 \\
    & 1024 & 76 & 73 & 78 & 76 \\
    & 1200 & 73 & 74 & 77 & 76 \\
    & 1600 & 70 & 72 & 65 & 68 \\
    & 2048 & 52 & 63 & 65 & 57 \\
    \bottomrule
    \end{tabular}
    }
  \caption{$100\times$COMET scores for \model{NLLB} (top) and \model{TowerBase} (bottom).} 
  \label{tab:res-base-comet}
\end{table}


\begin{table}
  \centering
      \scalebox{0.72}{

    \begin{tabular}{ll|rrrr}
    \toprule
     &       & count & mean       & min      & max        \\
    \midrule
    \multirow{2}{*}{\corpus{TED-full}}    
        & train & 1831  & 2915 /3515 & 56 /62   & 8706 /9706 \\
        & dev   & 19    & 2861 /3460 & 680 /867 & 6076 /7590 \\
    \midrule
    \multirow{2}{*}{\tedII} 
        & train & 15625 & 341 /411   & 3 /2     & 1022 /1460 \\
        & dev   & 160   & 339 /410   & 12 /17   & 959 /1203  \\
    \midrule
    \multirow{2}{*}{\tedI}   
        & train & 10582 & 504 /608   & 3 /1     & 1020 /1527 \\
        & dev   & 106   & 512 /620   & 42 /41   & 991 /1276 \\
    \bottomrule
    \end{tabular}
    }

  \caption{\rebuttal{Statistics of the TED talks training and dev sets. `count' denotes the number of parallel pseudo-documents. `mean', `min' and `max' represent the average, minimum and maximum lengths of English/French pseudo-documents respectively, in \model{NLLB} tokens.}
  }
  \label{tab:data-train-full}
\end{table}

\begin{table*}
  \centering
  \setlength{\tabcolsep}{3pt}
  
  \scalebox{0.7}{
\begin{tabular}{lr|rrrr|rrrr|rrrr|rrrr}
    \toprule
     & & \multicolumn{4}{c|}{2014} & \multicolumn{4}{c|}{2015} & \multicolumn{4}{c|}{2016} & \multicolumn{4}{c}{2017} \\
     & $l_{max}$ & count & min & max & mean & count & min & max & mean & count & min & max & mean & count & min & max & mean \\
     \midrule
     \parbox[t]{3mm}{\multirow{9}{*}{\rotatebox[origin=c]{90}{EN}}} &
     sent & 1335 & 2 & 112 & 23 & 1104 & 2 & 119 & 23 & 1185 & 1 & 151 & 24 & 1479 & 2 & 162 & 23 \\
    & 256 & 129 & 65 & 286 & 234 & 107 & 71 & 325 & 234 & 123 & 61 & 255 & 232 & 144 & 65 & 271 & 234 \\
    & 512 & 68 & 85 & 511 & 443 & 56 & 53 & 510 & 447 & 63 & 56 & 511 & 454 & 74 & 73 & 511 & 456 \\
    & 768 & 48 & 116 & 767 & 628 & 40 & 86 & 766 & 626 & 45 & 104 & 767 & 635 & 51 & 57 & 767 & 662 \\
    & 1024 & 37 & 83 & 1022 & 815 & 30 & 68 & 1023 & 835 & 35 & 115 & 1023 & 817 & 40 & 65 & 1023 & 844 \\
    & 1200 & 32 & 54 & 1218 & 942 & 26 & 71 & 1198 & 963 & 31 & 73 & 1216 & 922 & 34 & 125 & 1203 & 992 \\
    & 1600 & 26 & 135 & 1597 & 1160 & 24 & 114 & 1599 & 1043 & 23 & 191 & 1616 & 1243 & 27 & 229 & 1635 & 1250 \\
    & 2048 & 20 & 569 & 2091 & 1507 & 16 & 176 & 2072 & 1565 & 21 & 247 & 2046 & 1361 & 23 & 65 & 2045 & 1467 \\
    & doc & 15 & 995 & 4116 & 2010 & 12 & 1256 & 3359 & 2086 & 13 & 842 & 3366 & 2199 & 12 & 1909 & 3722 & 2812 \\

     \midrule
    \parbox[t]{3mm}{\multirow{9}{*}{\rotatebox[origin=c]{90}{FR}}} &
    sent & 1335 & 2 & 158 & 28 & 1104 & 2 & 145 & 27 & 1185 & 1 & 180 & 29 & 1479 & 2 & 211 & 27 \\
    & 256 & 129 & 80 & 380 & 295 & 107 & 85 & 355 & 282 & 123 & 69 & 345 & 276 & 144 & 78 & 375 & 275 \\
    & 512 & 68 & 106 & 717 & 559 & 56 & 62 & 679 & 540 & 63 & 70 & 672 & 539 & 74 & 83 & 737 & 535 \\
    & 768 & 48 & 142 & 1065 & 792 & 40 & 102 & 1009 & 755 & 45 & 112 & 985 & 755 & 51 & 68 & 1083 & 776 \\
    & 1024 & 37 & 100 & 1436 & 1027 & 30 & 64 & 1349 & 1007 & 35 & 125 & 1314 & 970 & 40 & 80 & 1431 & 990 \\
    & 1200 & 32 & 61 & 1641 & 1188 & 26 & 85 & 1577 & 1162 & 31 & 93 & 1511 & 1096 & 34 & 134 & 1714 & 1164 \\
    & 1600 & 26 & 173 & 2188 & 1462 & 24 & 156 & 2116 & 1259 & 23 & 209 & 2074 & 1477 & 27 & 279 & 2261 & 1466 \\
    & 2048 & 20 & 657 & 2613 & 1901 & 16 & 218 & 2602 & 1889 & 21 & 280 & 2626 & 1617 & 23 & 80 & 2714 & 1721 \\
    & doc & 15 & 1289 & 4983 & 2534 & 12 & 1609 & 4013 & 2518 & 13 & 1004 & 4179 & 2613 & 12 & 2473 & 4464 & 3299 \\
    \bottomrule
\end{tabular}}
  \caption{Statistics of the test sets based on talks from IWSLT \testset{tst2014}, \testset{tst2015}, \testset{tst2016} and \testset{tst2017} (see Section~\ref{ssec:exp-data}). \rebuttal{`count' refers to the number of parallel pseudo-documents. `mean', `min' and `max' denote the average, minimum and maximum lengths} of the source (i.e.\ English, top) or the reference (i.e.\ French, bottom) pseudo-documents. All lengths are counted in \model{NLLB} tokens. }
  \label{tab:data-testsets}
\end{table*}

\begin{table}
  \centering
    \scalebox{0.7}{
    \begin{tabular}{r|rrrrr}
        \toprule
         & \model{NLLB} & FT-U & Unif-U & FT-G & Unif-G \\
        \midrule
        sent & 0 & 0 & 0 & 0 & 0 \\
        256 & 557 & 6 & 6 & 6 & 6 \\
        512 & 1231 & 5 & 9 & 12 & 11 \\
        768 & 1618 & 6 & 10 & 250 & 280 \\
        1024 & 886 & 53 & 34 & 491 & 486 \\
        1200 & 1179 & 437 & 207 & 576 & 675 \\
        1600 & 352 & 465 & 657 & 789 & 840 \\
        2048 & 456 & 644 & 843 & 801 & 1089 \\
        \bottomrule
        \end{tabular}
    }

    \medskip
    \scalebox{0.7}{
    \begin{tabular}{r|rrrrr}
        \toprule
         & \model{Tower} & FT-U & Unif-U & FT-G & Unif-G \\
        \midrule
        sent & 0 & 0 & 0 & 0 & 0 \\
        256 & 79 & 3 & 8 & 2 & 3 \\
        512 & 58 & 2 & 2 & 3 & 2 \\
        768 & 65 & 4 & 3 & 3 & 5 \\
        1024 & 45 & 17 & 21 & 19 & 22 \\
        1200 & 107 & 19 & 17 & 13 & 19 \\
        1600 & 91 & 73 & 50 & 40 & 54 \\
        2048 & 151 & 94 & 84 & 66 & 98 \\
        \bottomrule
        \end{tabular}     
    }
  \caption{\edite{Number of empty alignments across all the $5,103$ sentences in our test sets for \model{NLLB} (top) and \model{TowerBase} (bottom) models.} } 
  \label{tab:stat-empty-align}
\end{table}

\subsection{Experimental Settings \label{ann:experiments}}
This section presents detailed experiment settings for fine-tuning \model{NLLB} and \model{TowerBase}.

For \model{NLLB}, we fine-tuned the pretrained model with learning rate $5\mathrm{e}{-4}$, 
$500$ warm-up steps, $4$ parallel pseudo-documents per batch and $32$ gradient accumulation steps. An early stopping criterion with a patience of $5$ epochs is also applied, according to the d-BLEU evaluated on the validation set. For inference on test sets, the beam size is set to $5$ and the batch size is set to $4$.

For \model{TowerBase}, We performed supervised fine-tuning using QLoRA \citep{dettmers2023qlora} and bfloat$16$.\footnote{
The prompt for fine-tuning is ``Translate the following text from English into French.\textbackslash nEnglish: SRC\textbackslash nFrench: TGT'', and the zero-shot prompt for the pretrained model \model{TowerBase} is ``English: SRC\textbackslash nFrench:''.  }
The batch size is $8$ with $2$ gradient accumulation steps. The learning rate is $2\mathrm{e}{-5}$ adjusted by a $cosine$ schedule, without warm-up steps nor packing. We fine-tuned the model for two epochs and saved checkpoints every $50$ steps in the second epoch. We then chose the checkpoint with the best d-BLEU on the validation set. Inference is performed without additional in-context examples, with bfloat$16$ and greedy search.

\subsection{Detailed Evaluation Results \label{ann:eval-score} }
The paired comparison and the complete BLEU and COMET scores for each test set are given in 
\Cref{tab:ttest-compare-system-base-ft,tab:res-nllb-ted-distill,tab:res-nllb-ted-unif,tab:res-tower-ted-distill,tab:res-tower-ted-unif}.
\paragraph{Document-level Fine-tuning}
Table~\ref{tab:ttest-compare-system-base-ft} reports average differences of automatic scores between fine-tuned MT systems and the corresponding pretrained models (\model{NLLB} or \model{TowerBase}), for varying test document lengths. ds-BLEUs are averaged over $52$ complete TED talks and COMET scores are averaged over $5,103$~sentences. Fine-tuning significantly improves over base conditions for all lengths, with larger increases for longer test texts, where the baseline scores were initially very poor. Both metrics yield consistent conclusions, except for the sentence-level assessment of \model{NLLB} fine-tuned on \tedI, which is slightly worse than the baseline according to ds-BLEU (-$0.8$), but for which COMET detects no difference. For \model{TowerBase}, DLFT is always beneficial.

\begin{table*}
  \centering
    \scalebox{0.8}{

\begin{tabular}{lr|rr|rr}
\toprule
 && \multicolumn{2}{c|}{ds-BLEU} & \multicolumn{2}{c}{COMET} \\
 && FT-U & FT-G & FT-U & FT-G \\
 \midrule
\parbox[t]{3mm}{\multirow{5}{*}{\rotatebox[origin=c]{90}{\model{NLLB}}}} &
sent & -0.8 \footnotesize{(0.00)} & -0.8 \footnotesize{(0.01)} & - & -0.3 \footnotesize{(0.01)} \\
& 256 & 7.6 \footnotesize{(0.00)} & 8.0 \footnotesize{(0.00)} & 13.0 \footnotesize{(0.00)} & 13.8 \footnotesize{(0.00)} \\
& 512 & 26.3 \footnotesize{(0.00)} & 27.0 \footnotesize{(0.00)} & 33.4 \footnotesize{(0.00)} & 33.8 \footnotesize{(0.00)} \\
& 768 & 33.5 \footnotesize{(0.00)} & 28.0 \footnotesize{(0.00)} & 39.0 \footnotesize{(0.00)} & 27.8 \footnotesize{(0.00)} \\
& 1024 & 33.5 \footnotesize{(0.00)} & 21.2 \footnotesize{(0.00)} & 41.9 \footnotesize{(0.00)} & 22.6 \footnotesize{(0.00)} \\
 \midrule
 \parbox[t]{3mm}{\multirow{8}{*}{\rotatebox[origin=c]{90}{\model{TowerBase}}}} &
sent & 2.4 \footnotesize{(0.00)} & 2.6 \footnotesize{(0.00)} & 0.7 \footnotesize{(0.00)} & 0.7 \footnotesize{(0.00)} \\
& 256 & 2.1 \footnotesize{(0.00)} & 1.6 \footnotesize{(0.00)} & 2.3 \footnotesize{(0.00)} & 2.3 \footnotesize{(0.00)} \\
& 512 & 2.1 \footnotesize{(0.00)} & 2.0 \footnotesize{(0.01)} & 2.4 \footnotesize{(0.00)} & 2.6 \footnotesize{(0.00)} \\
& 768 & 3.5 \footnotesize{(0.00)} & 3.2 \footnotesize{(0.00)} & 3.7 \footnotesize{(0.00)} & 3.7 \footnotesize{(0.00)} \\
& 1024 & 5.6 \footnotesize{(0.00)} & 5.0 \footnotesize{(0.00)} & 5.6 \footnotesize{(0.00)} & 5.5 \footnotesize{(0.00)} \\
& 1200 & 5.4 \footnotesize{(0.00)} & 5.1 \footnotesize{(0.00)} & 5.5 \footnotesize{(0.00)} & 5.4 \footnotesize{(0.00)} \\
& 1600 & 8.4 \footnotesize{(0.00)} & 8.3 \footnotesize{(0.00)} & 10.1 \footnotesize{(0.00)} & 10.3 \footnotesize{(0.00)} \\
& 2048 & 6.8 \footnotesize{(0.00)} & 6.1 \footnotesize{(0.00)} & 8.8 \footnotesize{(0.00)} & 7.9 \footnotesize{(0.00)} \\
\bottomrule
\end{tabular}
    
    }  
    \caption{Average difference (and p-values) in ds-BLEU or $100\times$COMET between fine-tuned models (FT) and the corresponding pretrained models \model{NLLB} (top) and \model{TowerBase} (bottom). U and G denote the corpora \tedI and \tedII respectively. \rebuttal{- for p-values $> 0.05$. }  Positive values indicate that the fine-tuned model improves over the baseline.
    } 
  \label{tab:ttest-compare-system-base-ft}
\end{table*}


\begin{table*} 
    \centering
    \begin{minipage}{0.5\textwidth}
        \scalebox{0.72}{
        \begin{tabular}{rl|rrrr}
        \toprule
         &  & 2014 & 2015 & 2016 & 2017 \\
         \midrule
         \multirow[c]{3}{*}{sent} & FT & 43.8 \footnotesize{(0.98)} & 44.0 \footnotesize{(0.99)} & 40.4 \footnotesize{(1.00)} & 41.3 \footnotesize{(1.00)} \\
         & Unif & 42.4 \footnotesize{(0.98)} & 42.5 \footnotesize{(0.99)} & 39.7 \footnotesize{(1.00)} & 40.2 \footnotesize{(1.00)} \\
         & SHAPE & 40.7 \footnotesize{(0.97)} & 41.5 \footnotesize{(0.98)} & 38.3 \footnotesize{(1.00)} & 39.4 \footnotesize{(1.00)} \\
         \midrule
        \multirow[c]{3}{*}{256} & FT & 43.5 \footnotesize{(0.98)} & 43.1 \footnotesize{(0.99)} & 40.4 \footnotesize{(1.00)} & 40.8 \footnotesize{(1.00)} \\
         & Unif & 43.9 \footnotesize{(0.98)} & 43.2 \footnotesize{(0.99)} & 39.7 \footnotesize{(1.00)} & 40.6 \footnotesize{(1.00)} \\
         & SHAPE & 43.3 \footnotesize{(0.98)} & 43.2 \footnotesize{(0.99)} & 39.8 \footnotesize{(1.00)} & 40.2 \footnotesize{(0.99)} \\
         \midrule
        \multirow[c]{3}{*}{512} & FT & 43.5 \footnotesize{(0.98)} & 43.4 \footnotesize{(0.98)} & 40.2 \footnotesize{(1.00)} & 40.4 \footnotesize{(1.00)} \\
         & Unif & 43.8 \footnotesize{(0.98)} & 43.8 \footnotesize{(0.99)} & 39.8 \footnotesize{(1.00)} & 41.0 \footnotesize{(1.00)} \\
         & SHAPE & 40.6 \footnotesize{(0.91)} & 40.5 \footnotesize{(0.92)} & 37.0 \footnotesize{(0.9)} & 40.4 \footnotesize{(0.98)} \\
         \midrule
        \multirow[c]{3}{*}{768} & FT & 36.6 \footnotesize{(0.87)} & 36.4 \footnotesize{(0.88)} & 35.3 \footnotesize{(0.92)} & 35.6 \footnotesize{(0.93)} \\
         & Unif & 41.8 \footnotesize{(0.95)} & 39.1 \footnotesize{(0.88)} & 38.1 \footnotesize{(0.95)} & 39.4 \footnotesize{(0.97)} \\
         & SHAPE & 34.0 \footnotesize{(0.75)} & 34.9 \footnotesize{(0.77)} & 33.8 \footnotesize{(0.84)} & 34.9 \footnotesize{(0.85)} \\
         \midrule
        \multirow[c]{3}{*}{1024} & FT & 28.6 \footnotesize{(0.70)} & 29.1 \footnotesize{(0.75)} & 28.9 \footnotesize{(0.80)} & 28.7 \footnotesize{(0.79)} \\
         & Unif & 36.1 \footnotesize{(0.81)} & 38.7 \footnotesize{(0.87)} & 34.9 \footnotesize{(0.87)} & 37.2 \footnotesize{(0.92)} \\
         & SHAPE & 32.4 \footnotesize{(0.73)} & 30.9 \footnotesize{(0.69)} & 30.8 \footnotesize{(0.75)} & 28.0 \footnotesize{(0.69)} \\
         \midrule
        \multirow[c]{3}{*}{1200} & FT & 25.2 \footnotesize{(0.64)} & 25.8 \footnotesize{(0.74)} & 24.1 \footnotesize{(0.71)} & 24.3 \footnotesize{(0.73)} \\
         & Unif & 34.6 \footnotesize{(0.80)} & 36.4 \footnotesize{(0.82)} & 30.0 \footnotesize{(0.74)} & 32.4 \footnotesize{(0.80)} \\
         & SHAPE & 27.2 \footnotesize{(0.61)} & 30.3 \footnotesize{(0.68)} & 24.6 \footnotesize{(0.64)} & 23.9 \footnotesize{(0.60)} \\
         \midrule
        \multirow[c]{3}{*}{1600} & FT & 18.2 \footnotesize{(0.53)} & 19.3 \footnotesize{(0.62)} & 19.2 \footnotesize{(0.62)} & 19.6 \footnotesize{(0.59)} \\
         & Unif & 25.5 \footnotesize{(0.59)} & 30.1 \footnotesize{(0.70)} & 26.9 \footnotesize{(0.68)} & 29.4 \footnotesize{(0.72)} \\
         & SHAPE & 22.0 \footnotesize{(0.50)} & 21.7 \footnotesize{(0.50)} & 22.1 \footnotesize{(0.56)} & 20.6 \footnotesize{(0.57)} \\
         \midrule
        \multirow[c]{3}{*}{2048} & FT & 15.4 \footnotesize{(0.49)} & 12.4 \footnotesize{(0.52)} & 16.7 \footnotesize{(0.58)} & 14.8 \footnotesize{(0.61)} \\
         & Unif & 22.0 \footnotesize{(0.51)} & 21.6 \footnotesize{(0.50)} & 24.3 \footnotesize{(0.60)} & 20.6 \footnotesize{(0.53)} \\
         & SHAPE & 18.7 \footnotesize{(0.43)} & 15.1 \footnotesize{(0.44)} & 14.6 \footnotesize{(0.38)} & 13.4 \footnotesize{(0.48)} \\
         \bottomrule
        \end{tabular}
        }
 
    \end{minipage}%
    \hspace{0.03\textwidth} 
    \begin{minipage}{0.45\textwidth}

    \scalebox{0.72}{

    \begin{tabular}{rl|rrrr}
    \toprule
     &  & 2014 & 2015 & 2016 & 2017 \\
     \midrule
    \multirow[c]{3}{*}{sent} & FT & 84 & 85 & 85 & 84 \\
     & Unif & 82 & 84 & 84 & 83 \\
     & SHAPE & 82 & 83 & 83 & 83 \\
     \midrule
    \multirow[c]{3}{*}{256} & FT & 81 & 82 & 82 & 81 \\
     & Unif & 81 & 82 & 82 & 81 \\
     & SHAPE & 80 & 82 & 81 & 80 \\
     \midrule
    \multirow[c]{3}{*}{512} & FT & 81 & 82 & 81 & 80 \\
     & Unif & 81 & 82 & 81 & 81 \\
     & SHAPE & 77 & 78 & 76 & 77 \\
     \midrule
    \multirow[c]{3}{*}{768} & FT & 69 & 69 & 70 & 69 \\
     & Unif & 74 & 75 & 74 & 73 \\
     & SHAPE & 68 & 69 & 68 & 68 \\
     \midrule
    \multirow[c]{3}{*}{1024} & FT & 58 & 59 & 60 & 58 \\
     & Unif & 67 & 65 & 67 & 67 \\
     & SHAPE & 60 & 62 & 61 & 59 \\
     \midrule
    \multirow[c]{3}{*}{1200} & FT & 56 & 56 & 55 & 53 \\
     & Unif & 61 & 65 & 62 & 60 \\
     & SHAPE & 55 & 59 & 55 & 54 \\
     \midrule
    \multirow[c]{3}{*}{1600} & FT & 50 & 49 & 49 & 49 \\
     & Unif & 55 & 58 & 54 & 56 \\
     & SHAPE & 51 & 52 & 50 & 48 \\
     \midrule
    \multirow[c]{3}{*}{2048} & FT & 47 & 42 & 48 & 45 \\
     & Unif & 51 & 49 & 51 & 48 \\
     & SHAPE & 46 & 43 & 47 & 44 \\
     \bottomrule
    \end{tabular}
}
    
    \caption{\edite{ds-BLEU (and brevity penalty)} (left) and $100\times$COMET (right) scores for \FTnllbII (FT), \UNIFnllbII (Unif), and \SHAPEnllbI (SHAPE) trained on \tedII with target max source document length $M=2048$.  }

    \label{tab:res-nllb-ted-distill}
    \end{minipage}
\end{table*}

\begin{table*} 
    \centering
    \begin{minipage}{0.5\textwidth}

    \scalebox{0.72}{
        \begin{tabular}{rl|rrrr}
        \toprule
         &  & 2014 & 2015 & 2016 & 2017 \\
         \midrule
         \multirow[c]{3}{*}{sent} & FT & 44.2 \footnotesize{(0.99)} & 43.5 \footnotesize{(0.99)} & 40.4 \footnotesize{(1.00)} & 41.4 \footnotesize{(1.00)} \\
         & Unif & 40.1 \footnotesize{(0.95)} & 40.4 \footnotesize{(0.97)} & 38.0 \footnotesize{(1.00)} & 38.1 \footnotesize{(0.99)} \\
         & SHAPE & 39.4 \footnotesize{(0.97)} & 40.0 \footnotesize{(0.98)} & 36.3 \footnotesize{(1.00)} & 37.8 \footnotesize{(0.99)} \\
         \midrule
        \multirow[c]{3}{*}{256} & FT & 43.2 \footnotesize{(0.98)} & 42.8 \footnotesize{(0.99)} & 39.7 \footnotesize{(1.00)} & 40.5 \footnotesize{(1.00)} \\
         & Unif & 42.8 \footnotesize{(0.98)} & 42.4 \footnotesize{(0.99)} & 39.5 \footnotesize{(1.00)} & 40.3 \footnotesize{(1.00)} \\
         & SHAPE & 42.0 \footnotesize{(0.97)} & 42.5 \footnotesize{(0.99)} & 39.6 \footnotesize{(1.00)} & 39.4 \footnotesize{(1.00)} \\
         \midrule
        \multirow[c]{3}{*}{512} & FT & 42.9 \footnotesize{(0.98)} & 43.1 \footnotesize{(0.99)} & 39.2 \footnotesize{(1.00)} & 39.4 \footnotesize{(1.00)} \\
         & Unif & 43.4 \footnotesize{(0.98)} & 43.0 \footnotesize{(0.99)} & 39.8 \footnotesize{(1.00)} & 40.5 \footnotesize{(1.00)} \\
         & SHAPE & 39.7 \footnotesize{(0.89)} & 41.1 \footnotesize{(0.94)} & 38.2 \footnotesize{(0.95)} & 39.0 \footnotesize{(0.98)} \\
         \midrule
        \multirow[c]{3}{*}{768} & FT & 43.5 \footnotesize{(0.98)} & 43.0 \footnotesize{(0.99)} & 39.4 \footnotesize{(1.00)} & 39.8 \footnotesize{(1.00)} \\
         & Unif & 44.0 \footnotesize{(0.98)} & 43.7 \footnotesize{(0.99)} & 40.4 \footnotesize{(1.00)} & 40.9 \footnotesize{(1.00)} \\
         & SHAPE & 39.6 \footnotesize{(0.88)} & 41.4 \footnotesize{(0.93)} & 37.4 \footnotesize{(0.93)} & 36.8 \footnotesize{(0.91)} \\
         \midrule
        \multirow[c]{3}{*}{1024} & FT & 42.6 \footnotesize{(0.96)} & 42.6 \footnotesize{(0.97)} & 39.2 \footnotesize{(1.00)} & 39.6 \footnotesize{(1.00)} \\
         & Unif & 42.6 \footnotesize{(0.96)} & 44.1 \footnotesize{(0.98)} & 39.6 \footnotesize{(1.00)} & 40.6 \footnotesize{(0.99)} \\
         & SHAPE & 40.3 \footnotesize{(0.91)} & 42.8 \footnotesize{(0.96)} & 36.8 \footnotesize{(0.92)} & 37.8 \footnotesize{(0.94)} \\
         \midrule
        \multirow[c]{3}{*}{1200} & FT & 38.3 \footnotesize{(0.88)} & 39.3 \footnotesize{(0.91)} & 36.3 \footnotesize{(0.92)} & 36.4 \footnotesize{(0.92)} \\
         & Unif & 39.5 \footnotesize{(0.89)} & 43.0 \footnotesize{(0.97)} & 38.0 \footnotesize{(0.95)} & 39.2 \footnotesize{(0.98)} \\
         & SHAPE & 36.9 \footnotesize{(0.84)} & 37.2 \footnotesize{(0.87)} & 32.6 \footnotesize{(0.82)} & 34.3 \footnotesize{(0.86)} \\
         \midrule
        \multirow[c]{3}{*}{1600} & FT & 31.5 \footnotesize{(0.77)} & 30.4 \footnotesize{(0.73)} & 30.9 \footnotesize{(0.83)} & 30.3 \footnotesize{(0.80)} \\
         & Unif & 31.5 \footnotesize{(0.72)} & 34.4 \footnotesize{(0.82)} & 34.2 \footnotesize{(0.88)} & 33.8 \footnotesize{(0.84)} \\
         & SHAPE & 28.0 \footnotesize{(0.68)} & 29.4 \footnotesize{(0.68)} & 31.1 \footnotesize{(0.79)} & 31.7 \footnotesize{(0.82)} \\
         \midrule
        \multirow[c]{3}{*}{2048} & FT & 27.4 \footnotesize{(0.69)} & 24.0 \footnotesize{(0.63)} & 26.7 \footnotesize{(0.75)} & 23.6 \footnotesize{(0.71)} \\
         & Unif & 30.2 \footnotesize{(0.68)} & 25.3 \footnotesize{(0.60)} & 31.5 \footnotesize{(0.79)} & 28.0 \footnotesize{(0.68)} \\
         & SHAPE & 26.1 \footnotesize{(0.63)} & 27.9 \footnotesize{(0.66)} & 26.8 \footnotesize{(0.69)} & 26.9 \footnotesize{(0.74)} \\
         \bottomrule
        \end{tabular}
    }

    \end{minipage}%
    \hspace{0.03\textwidth} 
    \begin{minipage}{0.45\textwidth}

    \scalebox{0.72}{
    \begin{tabular}{rl|rrrr}
    \toprule
     &  & 2014 & 2015 & 2016 & 2017 \\
     \midrule
    \multirow[c]{3}{*}{sent} & FT & 84 & 85 & 85 & 84 \\
     & Unif & 82 & 83 & 83 & 83 \\
     & SHAPE & 81 & 82 & 82 & 82 \\
     \midrule
    \multirow[c]{3}{*}{256} & FT & 80 & 81 & 81 & 81 \\
     & Unif & 80 & 81 & 81 & 81 \\
     & SHAPE & 79 & 81 & 81 & 80 \\
     \midrule
    \multirow[c]{3}{*}{512} & FT & 80 & 81 & 81 & 81 \\
     & Unif & 81 & 82 & 81 & 81 \\
     & SHAPE & 75 & 79 & 79 & 78 \\
     \midrule
    \multirow[c]{3}{*}{768} & FT & 80 & 81 & 81 & 81 \\
     & Unif & 81 & 82 & 82 & 81 \\
     & SHAPE & 75 & 78 & 75 & 75 \\
     \midrule
    \multirow[c]{3}{*}{1024} & FT & 78 & 79 & 78 & 78 \\
     & Unif & 80 & 81 & 81 & 80 \\
     & SHAPE & 74 & 78 & 75 & 73 \\
     \midrule
    \multirow[c]{3}{*}{1200} & FT & 71 & 73 & 70 & 69 \\
     & Unif & 76 & 79 & 76 & 75 \\
     & SHAPE & 68 & 70 & 69 & 66 \\
     \midrule
    \multirow[c]{3}{*}{1600} & FT & 61 & 60 & 61 & 60 \\
     & Unif & 62 & 61 & 61 & 64 \\
     & SHAPE & 57 & 59 & 62 & 59 \\
     \midrule
    \multirow[c]{3}{*}{2048} & FT & 57 & 53 & 57 & 54 \\
     & Unif & 59 & 57 & 57 & 54 \\
     & SHAPE & 51 & 52 & 57 & 53 \\
     \bottomrule
    \end{tabular}

}
    
    \caption{\edite{ds-BLEU (and brevity penalty)} (left) and $100\times$COMET (right) scores for \FTnllbI (FT) \UNIFnllbI (Unif), and \SHAPEnllbI (SHAPE) trained on \tedI with target max source document length $M=2048$.  }
    \label{tab:res-nllb-ted-unif}
    \end{minipage}
\end{table*}

\begin{table*} 
    \centering
    \begin{minipage}{0.5\textwidth}

        \scalebox{0.72}{

        \begin{tabular}{ll|rrrr}
        \toprule
         &  & 2014 & 2015 & 2016 & 2017 \\
         \midrule
        \multirow[c]{3}{*}{sent} & FT & 46.5 \footnotesize{(0.98)} & 45.1 \footnotesize{(0.99)} & 42.3 \footnotesize{(1.00)} & 41.0 \footnotesize{(1.00)} \\
         & Unif & 46.5 \footnotesize{(0.98)} & 45.0 \footnotesize{(0.99)} & 42.3 \footnotesize{(1.00)} & 41.1 \footnotesize{(1.00)} \\
         & SHAPE & 46.4 \footnotesize{(0.98)} & 45.2 \footnotesize{(0.99)} & 42.4 \footnotesize{(1.00)} & 41.2 \footnotesize{(1.00)} \\
         \midrule
        \multirow[c]{3}{*}{256} & FT & 44.6 \footnotesize{(0.98)} & 45.1 \footnotesize{(0.99)} & 42.3 \footnotesize{(1.00)} & 41.9 \footnotesize{(1.00)} \\
         & Unif & 44.5 \footnotesize{(0.99)} & 45.1 \footnotesize{(0.99)} & 42.2 \footnotesize{(1.00)} & 41.8 \footnotesize{(1.00)} \\
         & SHAPE & 46.2 \footnotesize{(0.98)} & 45.2 \footnotesize{(0.99)} & 42.4 \footnotesize{(1.00)} & 42.1 \footnotesize{(1.00)} \\
         \midrule
        \multirow[c]{3}{*}{512} & FT & 43.7 \footnotesize{(0.98)} & 45.0 \footnotesize{(1.00)} & 41.4 \footnotesize{(1.00)} & 41.6 \footnotesize{(1.00)} \\
         & Unif & 43.7 \footnotesize{(0.98)} & 44.8 \footnotesize{(1.00)} & 41.4 \footnotesize{(1.00)} & 41.4 \footnotesize{(1.00)} \\
         & SHAPE & 45.5 \footnotesize{(0.98)} & 45.1 \footnotesize{(0.99)} & 41.6 \footnotesize{(1.00)} & 41.6 \footnotesize{(1.00)} \\
         \midrule
        \multirow[c]{3}{*}{768} & FT & 44.0 \footnotesize{(0.98)} & 44.2 \footnotesize{(0.99)} & 40.3 \footnotesize{(1.00)} & 40.7 \footnotesize{(1.00)} \\
         & Unif & 44.0 \footnotesize{(0.98)} & 44.2 \footnotesize{(0.99)} & 40.2 \footnotesize{(1.00)} & 40.7 \footnotesize{(1.00)} \\
         & SHAPE & 45.6 \footnotesize{(0.98)} & 44.4 \footnotesize{(0.99)} & 41.3 \footnotesize{(1.00)} & 41.5 \footnotesize{(1.00)} \\
         \midrule
        \multirow[c]{3}{*}{1024} & FT & 42.7 \footnotesize{(0.98)} & 40.6 \footnotesize{(0.99)} & 38.9 \footnotesize{(1.00)} & 40.4 \footnotesize{(1.00)} \\
         & Unif & 42.2 \footnotesize{(0.97)} & 42.6 \footnotesize{(0.99)} & 39.4 \footnotesize{(1.00)} & 40.1 \footnotesize{(1.00)} \\
         & SHAPE & 44.5 \footnotesize{(0.98)} & 40.9 \footnotesize{(0.99)} & 39.2 \footnotesize{(1.00)} & 39.7 \footnotesize{(1.00)} \\
         \midrule
        \multirow[c]{3}{*}{1200} & FT & 42.6 \footnotesize{(0.98)} & 43.0 \footnotesize{(0.99)} & 38.9 \footnotesize{(1.00)} & 40.8 \footnotesize{(1.00)} \\
         & Unif & 42.5 \footnotesize{(0.98)} & 42.7 \footnotesize{(0.99)} & 39.3 \footnotesize{(1.00)} & 40.6 \footnotesize{(1.00)} \\
         & SHAPE & 44.0 \footnotesize{(0.98)} & 42.8 \footnotesize{(0.99)} & 39.3 \footnotesize{(1.00)} & 40.6 \footnotesize{(1.00)} \\
         \midrule
        \multirow[c]{3}{*}{1600} & FT & 42.0 \footnotesize{(0.97)} & 41.1 \footnotesize{(0.98)} & 37.0 \footnotesize{(1.00)} & 38.7 \footnotesize{(1.00)} \\
         & Unif & 42.0 \footnotesize{(0.97)} & 40.0 \footnotesize{(0.98)} & 37.5 \footnotesize{(1.00)} & 37.3 \footnotesize{(1.00)} \\
         & SHAPE & 42.2 \footnotesize{(0.97)} & 40.6 \footnotesize{(0.98)} & 38.5 \footnotesize{(1.00)} & 39.2 \footnotesize{(0.99)} \\
         \midrule
        \multirow[c]{3}{*}{2048} & FT & 33.0 \footnotesize{(0.99)} & 32.5 \footnotesize{(0.99)} & 31.6 \footnotesize{(1.00)} & 29.2 \footnotesize{(0.97)} \\
         & Unif & 33.1 \footnotesize{(0.99)} & 33.7 \footnotesize{(0.99)} & 32.2 \footnotesize{(0.98)} & 26.5 \footnotesize{(0.97)} \\
         & SHAPE & 34.1 \footnotesize{(0.97)} & 35.1 \footnotesize{(0.98)} & 32.1 \footnotesize{(1.00)} & 30.2 \footnotesize{(0.97)} \\
         \bottomrule
        \end{tabular}
    
    }
    \end{minipage}%
    \hspace{0.03\textwidth} 
    \begin{minipage}{0.45\textwidth}
    
    \scalebox{0.72}{
    \begin{tabular}{ll|rrrr}
    \toprule
     &  & 2014 & 2015 & 2016 & 2017 \\
     \midrule
     \multirow[c]{3}{*}{sent} & FT & 85 & 86 & 85 & 85 \\
     & Unif & 85 & 86 & 85 & 85 \\
     & SHAPE & 85 & 86 & 85 & 85 \\
     \midrule
    \multirow[c]{3}{*}{256} & FT & 83 & 84 & 83 & 83 \\
     & Unif & 83 & 84 & 83 & 83 \\
     & SHAPE & 83 & 84 & 83 & 83 \\
     \midrule
    \multirow[c]{3}{*}{512} & FT & 82 & 84 & 83 & 82 \\
     & Unif & 82 & 84 & 83 & 82 \\
     & SHAPE & 82 & 84 & 83 & 82 \\
     \midrule
    \multirow[c]{3}{*}{768} & FT & 82 & 83 & 83 & 82 \\
     & Unif & 82 & 83 & 83 & 82 \\
     & SHAPE & 82 & 84 & 83 & 82 \\
     \midrule
    \multirow[c]{3}{*}{1024} & FT & 81 & 81 & 82 & 81 \\
     & Unif & 81 & 82 & 83 & 81 \\
     & SHAPE & 81 & 81 & 83 & 81 \\
     \midrule
    \multirow[c]{3}{*}{1200} & FT & 78 & 82 & 81 & 81 \\
     & Unif & 78 & 82 & 81 & 81 \\
     & SHAPE & 80 & 82 & 82 & 81 \\
     \midrule
    \multirow[c]{3}{*}{1600} & FT & 80 & 79 & 79 & 80 \\
     & Unif & 79 & 78 & 79 & 78 \\
     & SHAPE & 79 & 79 & 80 & 79 \\
     \midrule
    \multirow[c]{3}{*}{2048} & FT & 68 & 69 & 68 & 64 \\
     & Unif & 69 & 70 & 70 & 62 \\
     & SHAPE & 65 & 73 & 70 & 68 \\

     \bottomrule
    \end{tabular}
    }
    \caption{\edite{ds-BLEU (and brevity penalty)} (left) and $100\times$COMET (right) scores for \FTtowerII (FT), \UNIFtowerII (Unif) and \SHAPEtowerII (SHAPE) trained on \tedII with target max source document length 2048 ($M = 4096$).}
    \label{tab:res-tower-ted-distill}
    \end{minipage}
\end{table*}

\begin{table*} 
    \centering
    \begin{minipage}{0.5\textwidth}

       \scalebox{0.72}{
       \begin{tabular}{rl|rrrr}
       \toprule
         &  & 2014 & 2015 & 2016 & 2017 \\
         \midrule
                 \multirow[c]{3}{*}{sent} & FT & 46.2 \footnotesize{(0.99)} & 45.1 \footnotesize{(0.99)} & 42.1 \footnotesize{(1.00)} & 40.8 \footnotesize{(1.00)} \\
         & Unif & 46.4 \footnotesize{(0.98)} & 45.1 \footnotesize{(0.99)} & 42.2 \footnotesize{(1.00)} & 40.9 \footnotesize{(1.00)} \\
         & SHAPE & 46.3 \footnotesize{(0.98)} & 45.2 \footnotesize{(0.99)} & 42.4 \footnotesize{(1.00)} & 41.0 \footnotesize{(1.00)} \\
         \midrule
        \multirow[c]{3}{*}{256} & FT & 46.3 \footnotesize{(0.98)} & 45.1 \footnotesize{(0.99)} & 42.3 \footnotesize{(1.00)} & 41.8 \footnotesize{(1.00)} \\
         & Unif & 44.5 \footnotesize{(0.98)} & 45.2 \footnotesize{(0.99)} & 42.3 \footnotesize{(1.00)} & 41.9 \footnotesize{(1.00)} \\
         & SHAPE & 46.0 \footnotesize{(0.98)} & 45.3 \footnotesize{(0.99)} & 42.4 \footnotesize{(1.00)} & 42.0 \footnotesize{(1.00)} \\
         \midrule
        \multirow[c]{3}{*}{512} & FT & 44.4 \footnotesize{(0.98)} & 44.9 \footnotesize{(0.99)} & 41.2 \footnotesize{(1.00)} & 41.6 \footnotesize{(1.00)} \\
         & Unif & 43.0 \footnotesize{(0.98)} & 45.0 \footnotesize{(0.99)} & 41.4 \footnotesize{(1.00)} & 41.6 \footnotesize{(1.00)} \\
         & SHAPE & 44.5 \footnotesize{(0.98)} & 45.0 \footnotesize{(0.99)} & 41.3 \footnotesize{(1.00)} & 41.6 \footnotesize{(1.00)} \\
         \midrule
        \multirow[c]{3}{*}{768} & FT & 44.1 \footnotesize{(0.98)} & 44.6 \footnotesize{(0.99)} & 41.1 \footnotesize{(1.00)} & 40.8 \footnotesize{(1.00)} \\
         & Unif & 43.2 \footnotesize{(0.99)} & 44.5 \footnotesize{(0.99)} & 41.4 \footnotesize{(1.00)} & 41.0 \footnotesize{(1.00)} \\
         & SHAPE & 44.2 \footnotesize{(0.99)} & 44.5 \footnotesize{(0.99)} & 41.2 \footnotesize{(1.00)} & 41.7 \footnotesize{(1.00)} \\
         \midrule
        \multirow[c]{3}{*}{1024} & FT & 43.9 \footnotesize{(0.97)} & 41.6 \footnotesize{(0.99)} & 39.6 \footnotesize{(1.00)} & 39.7 \footnotesize{(1.00)} \\
         & Unif & 42.7 \footnotesize{(0.98)} & 43.8 \footnotesize{(0.99)} & 39.7 \footnotesize{(1.00)} & 39.6 \footnotesize{(1.00)} \\
         & SHAPE & 44.1 \footnotesize{(0.98)} & 42.9 \footnotesize{(0.99)} & 39.8 \footnotesize{(1.00)} & 39.5 \footnotesize{(1.00)} \\
         \midrule
        \multirow[c]{3}{*}{1200} & FT & 43.2 \footnotesize{(0.97)} & 42.9 \footnotesize{(0.99)} & 39.5 \footnotesize{(1.00)} & 40.8 \footnotesize{(1.00)} \\
         & Unif & 43.2 \footnotesize{(0.98)} & 43.1 \footnotesize{(0.99)} & 38.2 \footnotesize{(1.00)} & 40.9 \footnotesize{(1.00)} \\
         & SHAPE & 43.8 \footnotesize{(0.98)} & 43.2 \footnotesize{(0.99)} & 39.2 \footnotesize{(1.00)} & 39.7 \footnotesize{(1.00)} \\
         \midrule
        \multirow[c]{3}{*}{1600} & FT & 41.6 \footnotesize{(0.95)} & 40.5 \footnotesize{(0.97)} & 37.6 \footnotesize{(1.00)} & 39.8 \footnotesize{(0.99)} \\
         & Unif & 41.1 \footnotesize{(0.97)} & 41.6 \footnotesize{(0.98)} & 36.5 \footnotesize{(1.00)} & 38.0 \footnotesize{(1.00)} \\
         & SHAPE & 42.6 \footnotesize{(0.96)} & 42.1 \footnotesize{(0.98)} & 37.3 \footnotesize{(1.00)} & 40.5 \footnotesize{(1.00)} \\
         \midrule
        \multirow[c]{3}{*}{2048} & FT & 34.4 \footnotesize{(0.96)} & 35.3 \footnotesize{(0.97)} & 31.2 \footnotesize{(0.99)} & 28.2 \footnotesize{(0.96)} \\
         & Unif & 34.6 \footnotesize{(0.97)} & 35.5 \footnotesize{(0.98)} & 30.2 \footnotesize{(1.00)} & 30.9 \footnotesize{(0.97)} \\
         & SHAPE & 35.2 \footnotesize{(0.97)} & 34.6 \footnotesize{(0.95)} & 31.9 \footnotesize{(0.99)} & 31.5 \footnotesize{(0.96)} \\
         \bottomrule
        \end{tabular}
    }

    \end{minipage}%
    \hspace{0.03\textwidth} 
    \begin{minipage}{0.45\textwidth}

        \scalebox{0.72}{
    
    \begin{tabular}{rl|rrrr}
    \toprule
     &  & 2014 & 2015 & 2016 & 2017 \\
     \midrule
     \multirow[c]{3}{*}{sent} & FT & 85 & 86 & 85 & 85 \\
     & Unif & 85 & 86 & 85 & 85 \\
     & SHAPE & 85 & 86 & 85 & 86 \\
     \midrule
    \multirow[c]{3}{*}{256} & FT & 83 & 84 & 83 & 83 \\
     & Unif & 83 & 84 & 83 & 82 \\
     & SHAPE & 83 & 84 & 83 & 82 \\
     \midrule
    \multirow[c]{3}{*}{512} & FT & 82 & 84 & 83 & 82 \\
     & Unif & 82 & 84 & 83 & 82 \\
     & SHAPE & 82 & 84 & 83 & 82 \\
     \midrule
    \multirow[c]{3}{*}{768} & FT & 82 & 83 & 83 & 82 \\
     & Unif & 82 & 84 & 83 & 82 \\
     & SHAPE & 82 & 83 & 83 & 82 \\
     \midrule
    \multirow[c]{3}{*}{1024} & FT & 81 & 81 & 83 & 81 \\
     & Unif & 81 & 83 & 83 & 81 \\
     & SHAPE & 81 & 82 & 83 & 81 \\
     \midrule
    \multirow[c]{3}{*}{1200} & FT & 80 & 82 & 81 & 81 \\
     & Unif & 80 & 82 & 81 & 81 \\
     & SHAPE & 80 & 82 & 81 & 80 \\
     \midrule
    \multirow[c]{3}{*}{1600} & FT & 79 & 79 & 80 & 80 \\
     & Unif & 80 & 80 & 78 & 79 \\
     & SHAPE & 79 & 81 & 78 & 80 \\
     \midrule
    \multirow[c]{3}{*}{2048} & FT & 68 & 72 & 69 & 64 \\
     & Unif & 70 & 72 & 69 & 67 \\
     & SHAPE & 70 & 73 & 70 & 68 \\
     \bottomrule
    \end{tabular}
    }
    
    \caption{\edite{ds-BLEU (and brevity penalty)} (left) and $100\times$COMET (right) for \FTtowerI (FT), \UNIFtowerI (Unif) and \SHAPEtowerI (SHAPE) trained on \tedI with target max source document length 2048 ($M = 4096$).}
    \label{tab:res-tower-ted-unif}
    \end{minipage}
\end{table*}

\begin{table*}
  \centering
  \setlength{\tabcolsep}{6pt}
    \scalebox{0.7}{
    \begin{tabular}{r|rr|rr|rrr}
    \toprule
     & \multicolumn{2}{c|}{\tedI} & \multicolumn{2}{c|}{\tedII} & FT & Unif & SHAPE \\
     & FT vs Unif & FT vs SHAPE & FT vs Unif & FT vs SHAPE & U vs G & U vs G & U vs G \\
     \midrule
    sent & -0.1 \footnotesize{(0.20)} & \textbf{-0.2 \footnotesize{(0.01)}} & -0.0 \footnotesize{(0.60)} & -0.1 \footnotesize{(0.18)} & \textbf{-0.1 \footnotesize{(0.05)}} & -0.1 \footnotesize{(0.19)} & -0.1 \footnotesize{(0.36)} \\
    256 & 0.5 \footnotesize{(0.32)} & -0.0 \footnotesize{(0.87)} & 0.1 \footnotesize{(0.22)} & -0.5 \footnotesize{(0.24)} & 0.5 \footnotesize{(0.32)} & 0.1 \footnotesize{(0.21)} & -0.1 \footnotesize{(0.48)} \\
    512 & 0.3 \footnotesize{(0.55)} & -0.1 \footnotesize{(0.10)} & 0.1 \footnotesize{(0.08)} & -0.6 \footnotesize{(0.23)} & 0.1 \footnotesize{(0.87)} & -0.1 \footnotesize{(0.82)} & -0.4 \footnotesize{(0.28)} \\
    768 & 0.2 \footnotesize{(0.46)} & -0.2 \footnotesize{(0.67)} & 0.0 \footnotesize{(0.84)} & \textbf{-0.9 \footnotesize{(0.05)}} & 0.3 \footnotesize{(0.11)} & 0.2 \footnotesize{(0.65)} & -0.4 \footnotesize{(0.19)} \\
    1024 & -0.2 \footnotesize{(0.82)} & -0.3 \footnotesize{(0.49)} & -0.4 \footnotesize{(0.44)} & -0.5 \footnotesize{(0.35)} & 0.6 \footnotesize{(0.29)} & 0.4 \footnotesize{(0.34)} & 0.5 \footnotesize{(0.38)} \\
    1200 & 0.2 \footnotesize{(0.44)} & 0.1 \footnotesize{(0.82)} & 0.0 \footnotesize{(0.70)} & -0.4 \footnotesize{(0.06)} & 0.3 \footnotesize{(0.24)} & 0.1 \footnotesize{(0.71)} & -0.2 \footnotesize{(0.50)} \\
    1600 & 0.6 \footnotesize{(0.45)} & \textbf{-0.7 \footnotesize{(0.01)}} & 0.4 \footnotesize{(0.58)} & -0.5 \footnotesize{(0.40)} & 0.1 \footnotesize{(0.84)} & 0.0 \footnotesize{(1.00)} & 0.4 \footnotesize{(0.49)} \\
    2048 & -0.5 \footnotesize{(0.61)} & -1.0 \footnotesize{(0.22)} & 0.2 \footnotesize{(0.84)} & -1.3 \footnotesize{(0.19)} & 0.7 \footnotesize{(0.46)} & 1.3 \footnotesize{(0.13)} & 0.4 \footnotesize{(0.44)} \\

     \midrule

     sent & -0.0 \footnotesize{(0.52)} & -0.0 \footnotesize{(0.93)} & 0.0 \footnotesize{(0.38)} & 0.0 \footnotesize{(0.60)} & -0.0 \footnotesize{(0.95)} & 0.0 \footnotesize{(0.24)} & 0.0 \footnotesize{(0.56)} \\
    256 & 0.0 \footnotesize{(0.60)} & 0.0 \footnotesize{(0.85)} & 0.0 \footnotesize{(0.73)} & -0.1 \footnotesize{(0.07)} & -0.0 \footnotesize{(0.63)} & -0.1 \footnotesize{(0.46)} & \textbf{-0.2 \footnotesize{(0.05)}} \\
    512 & -0.1 \footnotesize{(0.15)} & \textbf{-0.2 \footnotesize{(0.04)}} & 0.1 \footnotesize{(0.16)} & -0.1 \footnotesize{(0.27)} & -0.2 \footnotesize{(0.09)} & 0.0 \footnotesize{(0.76)} & -0.0 \footnotesize{(0.62)} \\
    768 & -0.1 \footnotesize{(0.29)} & 0.1 \footnotesize{(0.28)} & -0.1 \footnotesize{(0.34)} & \textbf{-0.3 \footnotesize{(0.00)}} & 0.0 \footnotesize{(0.87)} & 0.0 \footnotesize{(0.69)} & \textbf{-0.4 \footnotesize{(0.00)}} \\
    1024 & -0.6 \footnotesize{(0.22)} & -0.5 \footnotesize{(0.32)} & -0.3 \footnotesize{(0.46)} & -0.2 \footnotesize{(0.09)} & 0.1 \footnotesize{(0.45)} & 0.3 \footnotesize{(0.10)} & 0.4 \footnotesize{(0.42)} \\
    1200 & 0.1 \footnotesize{(0.69)} & 0.2 \footnotesize{(0.51)} & 0.1 \footnotesize{(0.50)} & -0.4 \footnotesize{(0.16)} & 0.1 \footnotesize{(0.61)} & 0.0 \footnotesize{(0.93)} & -0.4 \footnotesize{(0.12)} \\
    1600 & 0.2 \footnotesize{(0.69)} & -0.4 \footnotesize{(0.27)} & 0.6 \footnotesize{(0.37)} & 0.1 \footnotesize{(0.70)} & -0.2 \footnotesize{(0.62)} & 0.2 \footnotesize{(0.73)} & 0.4 \footnotesize{(0.49)} \\
    2048 & -0.9 \footnotesize{(0.40)} & -1.3 \footnotesize{(0.16)} & -0.2 \footnotesize{(0.84)} & -1.7 \footnotesize{(0.13)} & 0.9 \footnotesize{(0.50)} & 1.5 \footnotesize{(0.10)} & 0.4 \footnotesize{(0.47)} \\

    \bottomrule
    \end{tabular}
    }
    
    \caption{Average difference (and p-values) in \textbf{ds-BLEU} (top) evaluated on full TED talks and $100\times$\textbf{COMET} (bottom) evaluated on realigned sentences for \model{TowerBase}. \rebuttal{
    Left and middle: paired comparison between the original fine-tuning (FT), \posUnif and SHAPE on \tedI and \tedII respectively.
    Right: differences between fine-tuning on TED-U (U) and TED-G (G). A positive value indicates that in the comparison pair, the translation of the first item achieves higher scores than that of the second. Significant differences with p-values~$< 0.05$ are in bold.}
    } 
  \label{tab:ttest-compare-system-tower}
\end{table*}



\paragraph{Realignment Issues}
\edite{Since the COMET score is sentence-based, its computation requires a realignment between hypotheses and reference sentences in the Doc2Doc scenario. However, due to issues with long documents, translation hypotheses can be incomplete, resulting in empty alignment for some sentences. These untranslated sentences often occur in the final part of long documents. 
Table~\ref{tab:stat-empty-align} displays the statistics of empty alignments across all the $5,103$ sentences. This issue is more severe for \model{NLLB} models than \model{TowerBase} models, which is consistent with the poor BP reported in \Cref{tab:res-nllb-ted-distill,tab:res-nllb-ted-unif}.
}

\done{change all BLEU score to ds-BLEU at corpus level}
\done{check again the remarks of Paul and Rachel}
\end{document}